\documentclass{article}
\usepackage{ijcai26}

\usepackage{times}
\usepackage{soul}
\usepackage{url}
\usepackage[hidelinks]{hyperref}
\usepackage[utf8]{inputenc}
\usepackage[small]{caption}
\usepackage{graphicx}
\usepackage{amsmath}
\usepackage{amssymb}
\usepackage{amsthm}
\usepackage{booktabs}
\usepackage{paralist}
\usepackage{color}
\usepackage{pifont}
\usepackage{xcolor}

\usepackage{multirow}  
\usepackage{makecell}  

\title{InterLight: Leveraging Intrinsic Illumination Priors for Low-Light Image Enhancement}

\author{
    Ziqi Wang\textsuperscript{\rm 1}, 
    Xu Zhang\textsuperscript{\rm 1}\thanks{Corresponding author.}, 
    Laibin Chang\textsuperscript{\rm 1}, 
    Shi Chen\textsuperscript{\rm 2}, 
    Jiaqi Ma\textsuperscript{\rm 3}, 
    and Huan Zhang\textsuperscript{\rm 4}\thanks{Corresponding author.}
    \affiliations
    \textsuperscript{\rm 1}National Engineering Research Center for Multimedia Software, School of Computer Science, Wuhan University\\
    \textsuperscript{\rm 2}Department of Computer Science, University of Macau\\
    \textsuperscript{\rm 3}Department of Computer Vision, Mohamed bin Zayed University of Artificial Intelligence\\
    \textsuperscript{\rm 4}School of Information Engineering, Guangdong University of Technology
    \emails
    \{2023302112030, zhangx0802, changlb666\}@whu.edu.cn, chenshi@um.edu.mo, jiaqi.ma@mbzuai.ac.ae, huanzhang2021@gdut.edu.cn
}

\begin{document}

\maketitle

 \begin{abstract}

Low-Light Image Enhancement (LLIE) has long been a challenging problem in low-level vision, as insufficient illumination often leads to low contrast, detail loss, and noise. Recent studies show that deep learning-based Retinex theory can effectively decouple illumination and reflectance. However, existing methods frequently suffer from over-enhancement or color distortion, and often assume uniform noise or ideal lighting.
To address these limitations, we propose InterLight, a novel framework that systematically excavates and operationalizes intrinsic illumination priors for LLIE. 
Our core insight is that robust enhancement requires not just estimating illumination, but constructing an illumination-aware pipeline. 
We first inject sensor-level illumination-response priors via physics-guided augmentation, then represent the degradation through adaptive prompts conditioned on the scene's latent illumination state. This explicit representation directly guides a luminance-gated intrinsic memory mechanism to selectively compensate for information loss, prioritizing reconstruction in dark regions while preserving fidelity in bright ones. Finally, the entire process is regularized by a self-supervised consistency objective that distills illumination-invariant features. By deeply exploiting intrinsic illumination priors, our method achieves clearer textures and more visually coherent enhancement results. Extensive experiments across multiple benchmarks demonstrate the effectiveness of our approach. Code is available at: https://github.com/House-yuyu/InterLight.

\end{abstract}

\section{Introduction}

Low-Light Image Enhancement (LLIE)~\cite{retinex_03IJCV,zft_09TIP,LIME_16TIP,LOL,SRLLIE_18TIP,Zero-DCE,SCI_22CVPR,Retinexformer,SCI++,CWNet} is a fundamental challenge in computer vision, as images captured under insufficient illumination often suffer from severe noise, low contrast, color distortion, and structural detail loss. These degradations not only compromise visual quality but also hinder the performance of downstream tasks such as detection \cite{high_det_25ICCV}, segmentation \cite{high_Seg_25CVPR}, and autonomous driving \cite{high_AD_25NIPS}.

\begin{figure}[!t]
    \centering
    \includegraphics[width=1\linewidth]{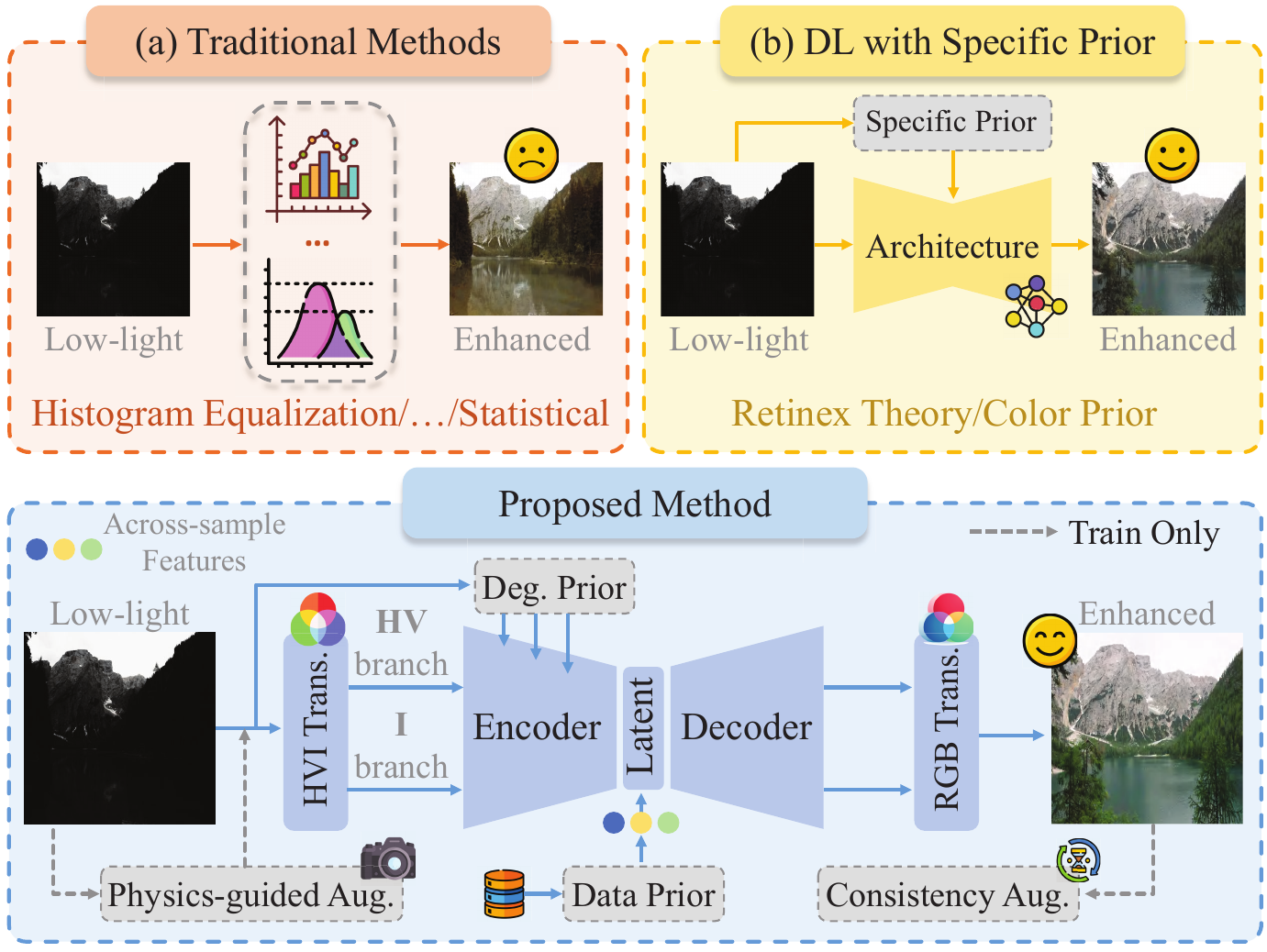}
    \caption{
    Comparison of {InterLight} with traditional and deep learning–based with specific prior LLIE methods.  
(a) Traditional approaches using handcrafted priors (e.g., histogram equalization, Retinex) often fail under non‑uniform illumination.
(b) Deep learning methods typically incorporate priors such as Retinex into new architectures to improve performance, but they still struggle with complex lighting and may introduce artifacts or color inconsistencies.
(c) In contrast, our method fully leverages intrinsic image priors to achieve more coherent and consistent enhancement.
    }
    \label{fig:motivation}
\end{figure}

As shown in Figure 1(a), early traditional LLIE methods primarily relied on handcrafted priors \cite{zft_09TIP,histograms_13TIP,naturalness_13TIP} to enhance brightness by expanding dynamic range and contrast, and on Retinex‑based models \cite{retinex_03IJCV,LIME_16TIP} that sought to separate illumination from reflectance.
Although effective in simple cases, these methods often struggled with complex degradations and exhibited limited generalization. With the advent of Deep Learning (DL), recent enhancement models \cite{CWNet,URetinexNet,CIDNet} have significantly improved restoration quality by learning mappings from low-light to normal-light images. 
For example, URetinexNet \cite{URetinexNet} unfolds the Retinex decomposition process into a learnable deep network, enabling more stable estimation of illumination and reflectance. CIDNet \cite{CIDNet} introduces a new HVI color space specifically designed for low-light enhancement, allowing the network to more effectively disentangle brightness and structural information. Although these methods can enhance images effectively, many of them still overlook the intrinsic illumination  priors inherent to low-light conditions, as illustrated in Figure 1(b).

To overcome these limitations, we introduce InterLight, a novel LLIE framework that systematically leverages intrinsic illumination priors. Unlike previous approaches that treat enhancement as generic image-to-image translation, InterLight builds an illumination-aware pipeline that injects sensor-level priors via physics-guided augmentation, represents degradations through adaptive prompts conditioned on latent illumination states, and employs a luminance-gated intrinsic memory to selectively compensate for information loss. The entire process is regularized by a self-supervised consistency objective, yielding natural colors, clearer textures, and visually coherent enhancements across diverse real-world scenarios.

In summary, our contributions are as follows:

\begin{itemize}
    \item 
We introduce InterLight, a new LLIE framework that systematically exploits intrinsic image priors. It constructs an illumination‑aware enhancement pipeline grounded in the physical and statistical properties of low‑light images.
    \item 
We propose ICDE, a physics‑guided augmentation strategy that simulates sensor‑level illumination responses while preserving structural fidelity. 
    \item 
We develop ADPG, which extracts sample‑specific degradation prior through a learnable degradation dictionary. The resulting prompt provides both global and spatially adaptive guidance for the chrominance branch via the PRFB.
    \item 
We design a LGIM that retrieves learned structural and textural patterns to compensate for information loss. It adaptively strengthens enhancement in dark regions while preserving fidelity in well‑lit areas.
\end{itemize}

\section{Related Work}
\subsection{Low-Light Image Enhancement}
Low-Light Image Enhancement (LLIE)~\cite{LOL,LightenDiff} aims to improve the quality and exposure of photographs captured in dark environments, producing dark-free images with natural color and proper illumination. Compared with traditional methods \cite{histograms_13TIP,naturalness_13TIP}, current deep learning-based approaches primarily focus on minimizing the reconstruction error between the enhanced output and the ground truth. Their main progress lies in refining network architectures or learning strategies tailored for LLIE. For example, Zero-DCE~\cite{Zero-DCE} estimates a reference curve via a deep network and performs zero-shot manner; PairLIE~\cite{PairLIE} adopts an unsupervised framework that learns adaptive priors from paired low-light images; Retinexformer~\cite{Retinexformer} predicts illumination to brighten the image and then restores degradations. Other architectures, including normalizing flow models~\cite{LLFlow}, diffusion models~\cite{JH_Diff_23TOG}, and Mamba-based models~\cite{Xu2025DSFormer}, have also been explored.
In contrast to these approaches, we advocate fully leveraging intrinsic illumination priors for LLIE to achieve more coherent and self-consistent enhancement results.

\subsection{Low-Light Image Enhancement with Intrinsic Priors} 
Early studies sought to improve LLIE performance by incorporating handcrafted priors \cite{zft_09TIP}, such as the Dark Channel Prior or histogram-based constraints. These optimization-based methods explicitly embraced the idea of decomposition and laid the foundation for leveraging illumination priors. Following this trend, recent deep learning-based Retinex \cite{LOL,DiffRetinex_25TPAMI,bai2025retinex} methods further extend this decomposition paradigm, yet their exploitation of illumination priors remains relatively shallow.
Another line of work incorporates high‑level visual priors \cite{ClearAIR,Perceive-IR,SMG_23CVPR,UniUIR}, where auxiliary cues such as semantic segmentation or depth are used to guide the enhancement process. The intuition is that semantic structure may help regularize low‑level restoration. However, these methods inevitably encounter a chicken‑and‑egg problem: reliable semantic or depth estimation itself requires sufficient visibility. Under complex real‑world low‑light conditions, such high‑level predictions often become unstable or noisy, which can in turn introduce incorrect guidance and lead to artifacts or structural inconsistencies in the enhanced results.
A third category of methods operates at the feature level \cite{xiao2025occlusion,LFPV_25ICCV,li2024object,11316813,Glare_24ECCV,chang2026color,UIC_zhu,di2025qmambabsr,he2023multispectral,Lookup_22CVPR}. Some approaches employ codebooks or lookup tables that store representative features, enabling each input to query and retrieve suitable references for the decoder.
In this work, we propose a hierarchical data–feature enhancement strategy fully exploiting intrinsic priors at both data-statistics and feature-representation levels. Besides, we explicitly incorporate degradation priors to guide adaptive restoration under diverse low-light conditions.

\begin{figure*}[!t]
    \centering
    \includegraphics[width=1\linewidth]{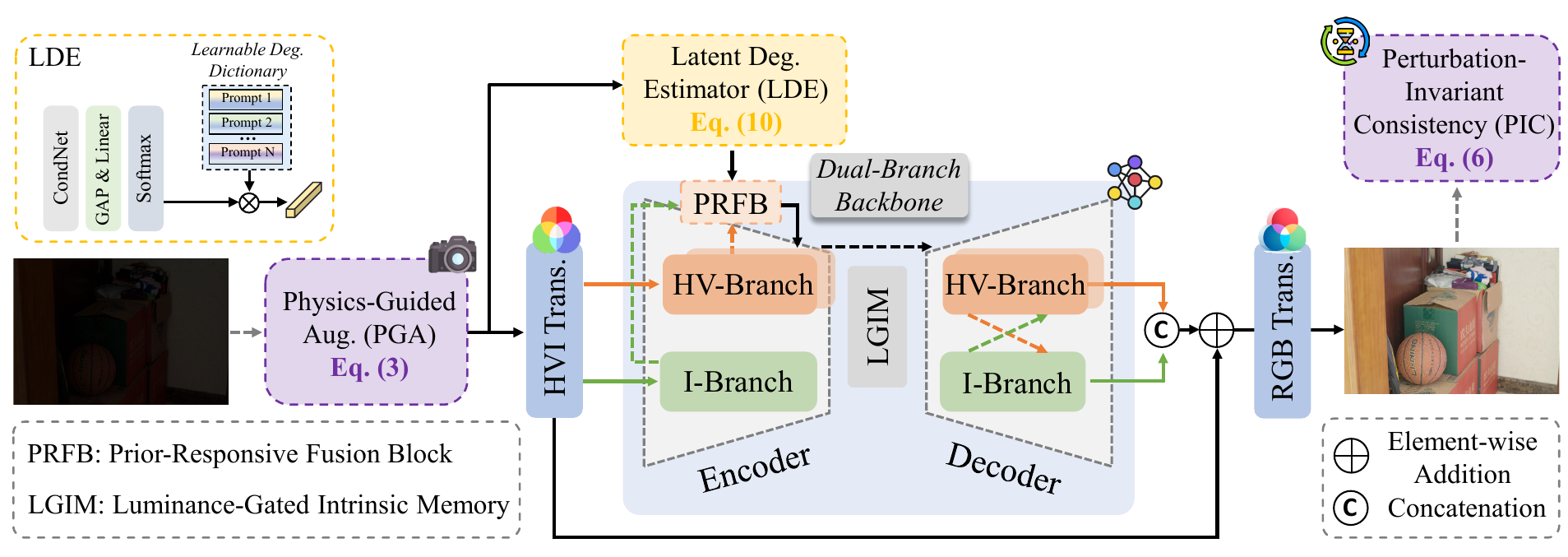}
    \caption{The overview of the proposed InterLight. The input is first augmented via PGA and then encoded by the LDE to produce an degradation prompt. After transforming the image into the HVI space, a dual‑branch network restores illumination and chrominance with prompt‑aware fusion. LGIM further compensates for information loss, and the final output is obtained through inverse HVI transformation with a residual connection.}
    \label{fig:overview}
\end{figure*}

\section{Method}
\label{sec:method}
In this paper, we propose InterLight, a novel LLIE framework that mines the intrinsic priors of low-light data without relying on external data or pre-trained priors. 
As illustrated in Figure~\ref{fig:overview}, our framework integrates three synergistic mechanisms to address data scarcity and feature degradation: (1) Intrinsic-Consistent Data Expansion to simulate physically plausible domain shifts and enforce structural consistency; (2) Adaptive Degradation Prior Generation to inject image-specific degradation contexts; and (3) Luminance-Gated Intrinsic Memory to retrieve restoration knowledge based on local feature.

\subsection{Overview}
\label{subsec:architecture}

Figure~\ref{fig:overview} illustrates the complete pipeline of InterLight. During training, the input $\mathbf{I}_{in}$ first undergoes Physics-Guided Augmentation (PGA) to simulate sensor variations. The augmented image is then processed by the Latent Degradation Estimator (LDE) to extract a global context prompt $\mathbf{p}$ encoding the degradation condition. Next, the input is decomposed into the HVI color space, yielding chrominance channels $(\mathcal{H}, \mathcal{V})$ and intensity channel $\mathcal{I}$.

The disentangled representation is processed through a dual-branch U-Net architecture. One branch restores illumination from $\mathcal{I}$, focusing on brightness recovery, while the other refines chrominance with prompt guidance via PRFB, preserving color fidelity. Both branches employ four-level encoder-decoders with Lightweight Cross-Attention (LCA) for mutual information exchange. At the bottleneck, the Luminance-Gated Intrinsic Memory (LGIM) retrieves learned patterns to compensate information loss, with stronger enhancement applied to darker regions.

Finally, the decoded features are fused through inverse HVI transformation with a global residual connection to produce the output $\mathbf{I}_{out}$. During training, Perturbation-Invariant Consistency (PIC) further enforces output stability via self-supervised augmentation. PGA and PIC are bypassed during inference.
\subsection{HVI Color Space}
\label{subsec:hvi_space}

To effectively decouple illumination and reflectance information, we adopt the Horizontal/Vertical-Intensity (HVI) color space introduced in~\cite{CIDNet}. Unlike standard color spaces, HVI incorporates a density-adaptive mechanism to handle the instability of chromaticity in low-light conditions.

Given an input image $\mathbf{I}_{in} \in \mathbb{R}^{H \times W \times 3}$, the intensity component is defined as the maximum value across channels, denoted as $\mathcal{P} = \max_c(\mathbf{I}_{in}^c)$. To mitigate noise in dark regions, a density function $\mathbf{C}_k$ is computed based on the intensity:
\begin{equation}
    \mathbf{C}_k = \left(\sin\left(\frac{\pi \mathcal{P}}{2}\right) + \epsilon\right)^k,
    \label{eq:Ck}
\end{equation}
where $k$ is a learnable parameter initialized to 0.2, and $\epsilon = 10^{-8}$ ensures numerical stability. As $\mathcal{P} \to 0$, $\mathbf{C}_k$ approaches zero, naturally suppressing unreliable color information.

Let $S$ and $H$ denote the standard saturation and normalized hue derived from the RGB-to-HSV conversion \cite{CIDNet}. The final HVI representation is formulated as:
\begin{equation} \label{eq:hvi}
    \mathcal{H} = \mathbf{C}_k \cdot S \cdot \cos(2\pi H), \quad
    \mathcal{V} = \mathbf{C}_k \cdot S \cdot \sin(2\pi H), \quad
    \mathcal{I} = \mathcal{P}.
\end{equation}
Here, $\mathcal{H}$ and $\mathcal{V}$ encode the chrominance information in a polar coordinate system modulated by $\mathbf{C}_k$, while $\mathcal{I}$ represents the illumination. The inverse transformation recovers the RGB image by normalizing $(\hat{\mathcal{H}}, \hat{\mathcal{V}})$ with $\mathbf{C}_k(\hat{\mathcal{I}})$ and applying the standard HSV-to-RGB conversion.

\subsection{Intrinsic-Consistent Data Expansion}
\label{subsec:icde}

To overcome data scarcity without external datasets, we introduce ICDE with two physics-aware strategies.

\paragraph{Physics-Guided Augmentation (PGA).}
Standard augmentation often violates low-light physics by amplifying noise in dark regions. PGA simulates sensor response variations through mild channel-wise Gamma correction $\gamma_c \sim \mathcal{U}(0.95, 1.05)$ for each channel $c \in \{R, G, B\}$, where $\mathcal{U}(a, b)$ denotes the uniform distribution over $[a, b]$. An information protection mechanism preserves dark regions:
\begin{equation}
    \mathbf{I}_{pga} = \alpha \cdot \mathbf{I}^{\gamma} + (1 - \alpha) \cdot \mathbf{I}, 
    \label{eq:PGA1}
\end{equation}
\begin{equation}
    \alpha = 3t^2 - 2t^3, \quad t = \min\left(1, \frac{\mathcal{P}}{\tau_d}\right),
    \label{eq:PGA2}
\end{equation}
where $\mathcal{P} = \max_c(\mathbf{I}_c)$ denotes the per-pixel maximum intensity across RGB channels, and $\tau_d = 0.05$ is the dark region threshold. The smoothstep function $\alpha \in [0, 1]$ ensures augmentation only affects regions with valid photon information ($\alpha \to 1$ when bright, $\alpha \to 0$ when dark), thereby avoiding non-physical artifacts in noise-dominated areas.

\paragraph{Perturbation-Invariant Consistency (PIC).}
Given an enhanced output $\mathbf{I}_{e}$, we generate weak and strong augmented views:
\begin{equation}
    \mathbf{I}_w = \text{CenterCrop}_{s}(\mathbf{I}_{e}), \quad
    \mathbf{I}_s = \mathcal{G}_{\sigma}(\mathbf{I}_w),
    \label{eq:pic_views}
\end{equation}
where $s=16$ is the crop size and $\mathcal{G}_{\sigma}$ denotes Gaussian blur with kernel size ranging from 9 to 21 and standard deviation $\sigma \sim \mathcal{U}(0.1, 5)$. The consistency loss minimizes the MSE distance between views:
\begin{equation}
    \mathcal{L}_{consistency} = \beta(t) \cdot \|\mathbf{I}_w - \mathbf{I}_s\|_2^2,
    \label{eq:consistency_loss}
\end{equation}
where the weight $\beta(t)$ follows a cosine decay schedule to prevent over-smoothing as training converges:
\begin{equation}
    \beta(t) = \frac{\beta_0}{2} \left( \cos\left(\frac{\pi t}{T}\right) + 1 \right),
    \label{eq:pic_decay}
\end{equation}
with $\beta_0 = 0.1$ being the initial weight and $T$ the total training steps. This encourages learning of scale- and blur-invariant features intrinsic to the data.

\subsection{Adaptive Degradation Prior Generation}
\label{subsec:adpg}

While ICDE addresses data diversity, global degradation levels vary significantly across samples. The ADPG module extracts and injects degradation priors through two components.

\paragraph{Latent Degradation Estimator (LDE).}
Given the PGA output $\mathbf{I}_{pga} \in \mathbb{R}^{3 \times H \times W}$, LDE first extracts a compact global feature vector using a lightweight condition network $\mathcal{C}$ (a stack of strided convolutions followed by $1 \times 1$ convolutions):
\begin{equation}
    \mathbf{z} = \text{GAP}\left(\mathcal{C}(\mathbf{I}_{pga})\right) \in \mathbb{R}^{d_z},
    \label{eq:lde_feat}
\end{equation}
where GAP denotes global average pooling and $d_z = 32$ is the latent feature dimension. We maintain a learnable degradation dictionary $\mathbf{D} = [\mathbf{d}_1, \ldots, \mathbf{d}_K]^\top \in \mathbb{R}^{K \times d_p}$, where each vector $\mathbf{d}_k \in \mathbb{R}^{d_p}$ represents a prototypical degradation pattern. The coefficients over this dictionary are computed via softmax normalization:
\begin{equation}
    \boldsymbol{\alpha} = \text{Softmax}\left(\mathbf{W}_\alpha \mathbf{z}\right) \in \mathbb{R}^{K},
    \label{eq:lde_coef}
\end{equation}
where $\mathbf{W}_\alpha \in \mathbb{R}^{K \times d_z}$ is a learnable projection matrix and $K = 32$ is the number of dictionary atoms. The context prompt $\mathbf{p}$ is obtained as a weighted combination of dictionary vectors:
\begin{equation}
    \mathbf{p} = \phi\left(\boldsymbol{\alpha}^\top \mathbf{D}\right) = \phi\left(\sum_{k=1}^{K} \alpha_k \mathbf{d}_k\right) \in \mathbb{R}^{d_p},
    \label{eq:lde_prompt}
\end{equation}
where $\phi(\cdot)$ denotes the GELU activation function and $d_p = 512$ is the prompt dimension. The dictionary vectors $\{\mathbf{d}_k\}$ are jointly learned during training, discovering representative degradation patterns from data without manual categorization.

\subsubsection{Prior-Responsive Fusion Block (PRFB)}

PRFB replaces standard cross-attention in the HV-branch encoder, modulating features using prompt $\mathbf{p}$. Specifically, $\mathbf{p}$ is first projected to channel-wise scale and shift parameters that adjust global feature statistics:
\begin{equation}
    \boldsymbol{\gamma} = \sigma\left(\mathbf{W}_\gamma \mathbf{p}\right) \in \mathbb{R}^{C}, \quad
    \boldsymbol{\beta} = \mathbf{W}_\beta \mathbf{p} \in \mathbb{R}^{C},
    \label{eq:prfb_affine_params}
\end{equation}
where $\sigma(\cdot)$ is the sigmoid function ensuring $\boldsymbol{\gamma} \in (0, 1)$, $C$ denotes the feature channel dimension, and $\mathbf{W}_\gamma, \mathbf{W}_\beta \in \mathbb{R}^{C \times d_p}$ are learnable linear projections. The input feature $\mathbf{X} \in \mathbb{R}^{C \times H \times W}$ is then modulated as:
\begin{equation}
    \mathbf{X}' = \mathbf{X} \odot (1 + \boldsymbol{\gamma}) + \boldsymbol{\beta},
    \label{eq:prfb_modulation}
\end{equation}
where $\odot$ denotes channel-wise multiplication with spatial broadcasting.

To enable spatially-adaptive guidance, the prompt is projected to a spatial feature map and upsampled:
\begin{equation}
\begin{split}
    \mathbf{P}_s &= \text{Upsample}_{H \times W}(\mathbf{U}) \in \mathbb{R}^{C \times H \times W}, \\
    \mathbf{U} &= \text{Conv}\left(\text{Reshape}_{h \times w}\left(\mathbf{W}_s \mathbf{p}\right)\right),
\end{split}
\label{eq:prfb_spatial}
\end{equation}
where $\mathbf{W}_s \in \mathbb{R}^{(C \cdot h \cdot w) \times d_p}$ projects the prompt to spatial dimensions, and $h, w$ are level-dependent spatial sizes (16, 8, 4 for shallow, middle, deep layers respectively). A learned gate $\mathbf{G}$ balances degradation guidance versus local content:
\begin{equation}
    \mathbf{G} = \sigma\left(\text{Conv}_{1\times1}\left(\text{DWConv}\left([\mathbf{q}_1; \mathbf{q}_2]\right)\right)\right) \in \mathbb{R}^{C \times H \times W},
    \label{eq:prfb_gate}
\end{equation}
where $\mathbf{q}_1 = \text{Conv}_{1\times1}(\mathbf{P}_s)$, $\mathbf{q}_2 = \text{Conv}_{1\times1}(\mathbf{X}')$ are query projections, $[\cdot; \cdot]$ denotes channel-wise concatenation, and DWConv denotes depth-wise convolution. The query for cross-attention is formulated as:
\begin{equation}
    \mathbf{Q} = \text{DWConv}\left(\mathbf{G} \odot \mathbf{q}_1 + (1 - \mathbf{G}) \odot \mathbf{q}_2\right).
    \label{eq:prfb_query}
\end{equation}

The cross-attention with the I-branch feature $\mathbf{Y} \in \mathbb{R}^{C \times H \times W}$ follows normalized attention:
\begin{equation}
    \mathbf{K}, \mathbf{V} = \text{Split}\left(\text{DWConv}\left(\text{Conv}_{1\times1}(\mathbf{Y})\right)\right),
    \label{eq:prfb_kv}
\end{equation}
\begin{equation}
    \text{Attn}(\mathbf{Q}, \mathbf{K}, \mathbf{V}) = \text{Softmax}\left(\frac{\bar{\mathbf{Q}}\bar{\mathbf{K}}^\top}{\tau_a}\right)\mathbf{V}.
    \label{eq:prfb_attn}
\end{equation}

The final output combines the attention result with a feed-forward network (FFN):
\begin{equation}
    \mathbf{X}_{out} = \mathbf{X} + \text{Attn}(\mathbf{Q}, \mathbf{K}, \mathbf{V}) + \text{FFN}(\mathbf{X}).
    \label{eq:prfb_output}
\end{equation}

This design enables the network to adapt its restoration behavior based on the estimated global illumination condition, with the gate $\mathbf{G}$ learning to emphasize prompt guidance in severely degraded regions while preserving local content in well-lit areas.

\subsection{Luminance-Gated Intrinsic Memory}
\label{subsec:lgim}

Low-light images often suffer from information loss that local convolutions cannot recover. LGIM builds an internal knowledge base to compensate.

Specifically, LGIM maintains $L$ learnable global vectors $\{\mathbf{m}_l^v\}_{l=1}^{L} \in \mathbb{R}^{C}$ designed to capture channel statistics alongside local patches $\{\mathbf{m}_l^p\}_{l=1}^{L} \in \mathbb{R}^{C \times r \times r}$ with patch size $r$ to capture textures. These entries are dynamically updated during training through sample-wise incorporation and mutual propagation. Given an input feature $\mathbf{F}_{in} \in \mathbb{R}^{C \times H \times W}$, the system queries this memory bank via attention mechanisms to retrieve relevant patterns at both vector and patch granularities, producing the retrieved memory feature denoted as $\mathbf{F}_{mem} \in \mathbb{R}^{C \times H \times W}$. Based on the critical observation that bright regions usually contain sufficient information whereas dark regions require aggressive hallucination, we implement an intensity-inverse modulation strategy. Unlike standard memory banks that execute spatially uniform feature retrieval, our LGIM adaptively bypasses memory intervention in high-fidelity regions while aggressively compensating for structural loss in dark areas. The final feature fusion is thus formulated as:
\begin{equation}
    \mathbf{F}_{out} = \mathbf{F}_{in} + \lambda \cdot (1 + \sigma(\eta)(1 - g)) \cdot \mathbf{F}_{mem},
    \label{eq:lgim}
\end{equation}
where $g = \sigma(\text{MLP}(\text{GAP}(\mathbf{F}_{in}))) \in [0,1]$ is a learned gate indicating local brightness estimated from input feature statistics, $\lambda$ is a learnable fusion scale, $\eta$ is a learnable adaptive strength parameter, and $\sigma(\cdot)$ denotes the sigmoid function. When $g \to 0$ (dark), the adaptive gain $(1 + \sigma(\eta)(1 - g))$ increases; when $g \to 1$ (bright), it decreases toward 1. This ensures aggressive memory retrieval for degraded dark regions while preserving fidelity in well-lit areas.

The I-branch uses stronger fusion ($\lambda_{init}=1.2$) with an additional learnable brightness bias $\mathbf{b} \in \mathbb{R}^{C}$, while the HV-branch uses conservative fusion ($\lambda_{init}=0.8$) with I-branch features as brightness reference for cross-branch guidance.

\subsection{Training Objectives}
\label{subsec:loss}

The network is supervised by a compound loss in both RGB and HVI domains:
\begin{equation}
    \mathcal{L}_{rec} = \mathcal{L}_{L1} + \mathcal{L}_{ssim} + \mathcal{L}_{edge} + \mu_p \mathcal{L}_{perc},
    \label{eq:loss_rec}
\end{equation}
combining pixel-wise L1 loss, structural SSIM loss, Laplacian edge loss, and VGG-based perceptual loss. The HVI-domain loss (weighted by $\mu_{hvi}$) ensures accurate reconstruction of both illumination and reflectance components:
\begin{equation}
    \mathcal{L}_{total} = \mathcal{L}_{rec}^{RGB} + \mu_{hvi} \mathcal{L}_{rec}^{HVI} + \mathcal{L}_{consistency}.
    \label{eq:loss_total}
\end{equation}

A dual-path training strategy supervises both baseline output $\hat{\mathbf{I}}_{base}$ (without LGIM) and memory-enhanced output $\hat{\mathbf{I}}_{mem}$:
\begin{equation}
    \mathcal{L}_{dual} = \mathcal{L}_{total}(\hat{\mathbf{I}}_{base}) + \lambda_{lgim} \mathcal{L}_{total}(\hat{\mathbf{I}}_{mem}),
    \label{eq:loss_dual}
\end{equation}
ensuring robust baseline performance while enabling complementary memory-based enhancement.


\section{Experiments}
\label{sec:experiments}

\subsection{Experimental Setup} 
\paragraph{Datasets and Metrics.}
Following established protocols in low-light image enhancement, we conduct comprehensive evaluations across multiple benchmarks. We adopt the LOL-v1 \cite{LOL}, LOL-v2 \cite{LOLV2} , SICE \cite{SICE} (including the Mix and Grad test sets, SID (Sony-Total-Dark) \cite{SID}, and LSRW-Huawei \cite{LSRW} datasets. We employ widely adopted metrics, including Peak Signal-to-Noise Ratio (PSNR) and Structural Similarity Index (SSIM) \cite{SSIM}, to quantify restoration fidelity. Higher scores indicate better performance.

\paragraph{Implementation Details.}

The training process is optimized using the Adam~\cite{Adam} optimizer with $\beta_1=0.9$ and $\beta_2=0.999$. The initial learning rate is set to $2 \times 10^{-4}$ and decays following a cosine annealing schedule. We train the model for 1500 epochs with a batch size of 8. Notably, InterLight consistently converges across four fundamentally distinct datasets (LOL, SID, SICE, LSRW) using this unified optimization setting, without requiring any per-dataset initialization tuning, thereby verifying its excellent optimization stability. The input images are randomly cropped to $256 \times 256$ patches and augmented with random horizontal and vertical flips.
The dual-branch encoder-decoder follows a four-level U-Net architecture with channel dimensions $[36, 36, 72, 144]$. All experiments are conducted on NVIDIA RTX 4090 GPUs using PyTorch. 

For the {ADPG} module, we employ a degradation dictionary with $N_a=32$ and prompt dimension $d_p=512$. The spatial prior resolution in {PRFB} is set to $\{16, 8, 4\}$ for the three encoder levels. For the {LGIM} module, we maintain $L=16$ memory entries with patch size $k=4$, and initialize the fusion scales to $\lambda_I=1.2$ for the I-branch and $\lambda_{HV}=0.8$ for the HV-branch.
The loss coefficients are set as follows: HVI-domain weight $\mu_{hvi}=0.5$, perceptual loss weight $\mu_p=0.1$, and LGIM dual-path weight $\lambda_{lgim}=1.0$. The PIC consistency loss weight $\beta_0$ is initialized to 0.1 and decays via cosine annealing.
For PGA, we apply channel-wise Gamma correction with $\gamma \sim \mathcal{U}(0.95, 1.05)$ at probability $p=0.3$. The dark protection threshold is set to $\tau=0.05$.


\begin{table*}[!htb]
    \centering
    \renewcommand{\arraystretch}{1.1} 

    \resizebox{\textwidth}{!}{
    \begin{tabular}{l|c|c|cc|cc|cc|cc}
        \toprule[1pt]
        
        \multirow{2}{*}{\textbf{Methods}} & \multirow{2}{*}{\textbf{Venue}} & \multirow{2}{*}{\textbf{Color Space}} & \multicolumn{2}{c|}{\textbf{Complexity}} & \multicolumn{2}{c|}{\textbf{LOL-v1}} & \multicolumn{2}{c|}{\textbf{LOL-v2-Real}} & \multicolumn{2}{c}{\textbf{LOL-v2-Syn}}\\

        \cline{4-11}

        ~ & ~ & ~ & Params/M & FLOPs/G & PSNR$\uparrow$ & SSIM$\uparrow$ & PSNR$\uparrow$ & SSIM$\uparrow$ & PSNR$\uparrow$ & SSIM$\uparrow$ \\
        
        \midrule
        
        RetinexNet \cite{LOL} & BMVC'18 & Retinex & 0.84 & 584.47 & 18.92 & 0.427 & 16.10 & 0.401 & 17.14 & 0.762 \\
        
        KinD \cite{KinD} & MM'19 & Retinex & 8.02 & 34.99 & 23.02 & 0.843 & 17.54 & 0.669 & 18.32 & 0.796 \\
        
        ZeroDCE \cite{Zero-DCE} & CVPR'20 & RGB & 0.075 & 4.83 & 21.88 & 0.640 & 16.06 & 0.580 & 17.71 & 0.815 \\
        
        RUAS \cite{RUAS} & CVPR'21 & Retinex & 0.003 & 0.83 & 18.65 & 0.518 & 15.33 & 0.488 & 13.77 & 0.638 \\
        
        EnlightenGAN \cite{EnGAN} & TIP'21 & RGB & 114.35 & 61.01 & 20.00 & 0.691 & 18.23 & 0.617 & 16.57 & 0.734 \\
        
        SNR-Aware \cite{SNR-Aware} & CVPR'22 & SNR+RGB & 4.01 & 26.35 & 24.61 & 0.842 & 21.48 & 0.849 & 24.14 & 0.928 \\
        
        Bread \cite{Bread} & IJCV'23 & YCbCr & 2.02 & 19.85 & 22.92 & 0.836 & 20.83 & 0.847 & 17.63 & 0.919 \\
        
        PairLIE \cite{PairLIE} & CVPR'23 & Retinex & 0.33 & 20.81 & 23.53 & 0.755 & 19.89 & 0.778 & 19.07 & 0.794 \\
        
        LLFormer \cite{LLFormer} & AAAI'23 & RGB & 24.55 & 22.52 & 23.65 & 0.816 & 20.06 & 0.792 & 24.04 & 0.909 \\
        
        Retinexformer \cite{Retinexformer} & ICCV'23 & Retinex & 1.53 & 15.85 & \textcolor{red}{25.16} & 0.845 & 22.79 & 0.840 & 25.67 & 0.930 \\
        
        LightenDiff \cite{LightenDiff} & ECCV'24 & RGB & 26.54 & 2257.42 & 23.62 & 0.829 & 22.88 & 0.855 & 21.58 & 0.869 \\
        
        CWNet \cite{CWNet} & ICCV'25 & RGB & 1.23 & 11.3 & 23.60 & 0.850 & 21.65 & 0.860 & 25.50 & \textcolor{blue}{0.936} \\

        
        CIDNet \cite{CIDNet} & CVPR'25 & HVI & 1.88 & 7.57 & 23.81 & \textcolor{blue}{0.857} &  \textcolor{blue}{23.90} & 0.865 & \textcolor{blue}{25.71} & \textcolor{red}{0.942} \\
        
        \textbf{Ours} & - & HVI & 10.91 & 8.41 & \textcolor{blue}{24.78} & \textcolor{red}{0.862} & \textcolor{red}{24.06} & \textcolor{red}{0.866} & \textcolor{red}{25.73} & 0.935 \\
        
        \bottomrule[1pt]
        
    \end{tabular}
    }
        \caption{Quantitative results of PSNR$\uparrow$ and SSIM$\uparrow$ on the LOL (v1 and v2) datasets. The best performance is in \textcolor{red}{red} color and the second best is in \textcolor{blue}{blue} color.}
    \label{tab:table-LOL}
\end{table*}

\begin{figure*}[!htb]
\centering
\newcommand{\eightcolwidth}{0.123\linewidth} 

\begin{minipage}[t]{\eightcolwidth}
    \centering
    \vspace{3pt}
    \centerline{\includegraphics[width=\textwidth]{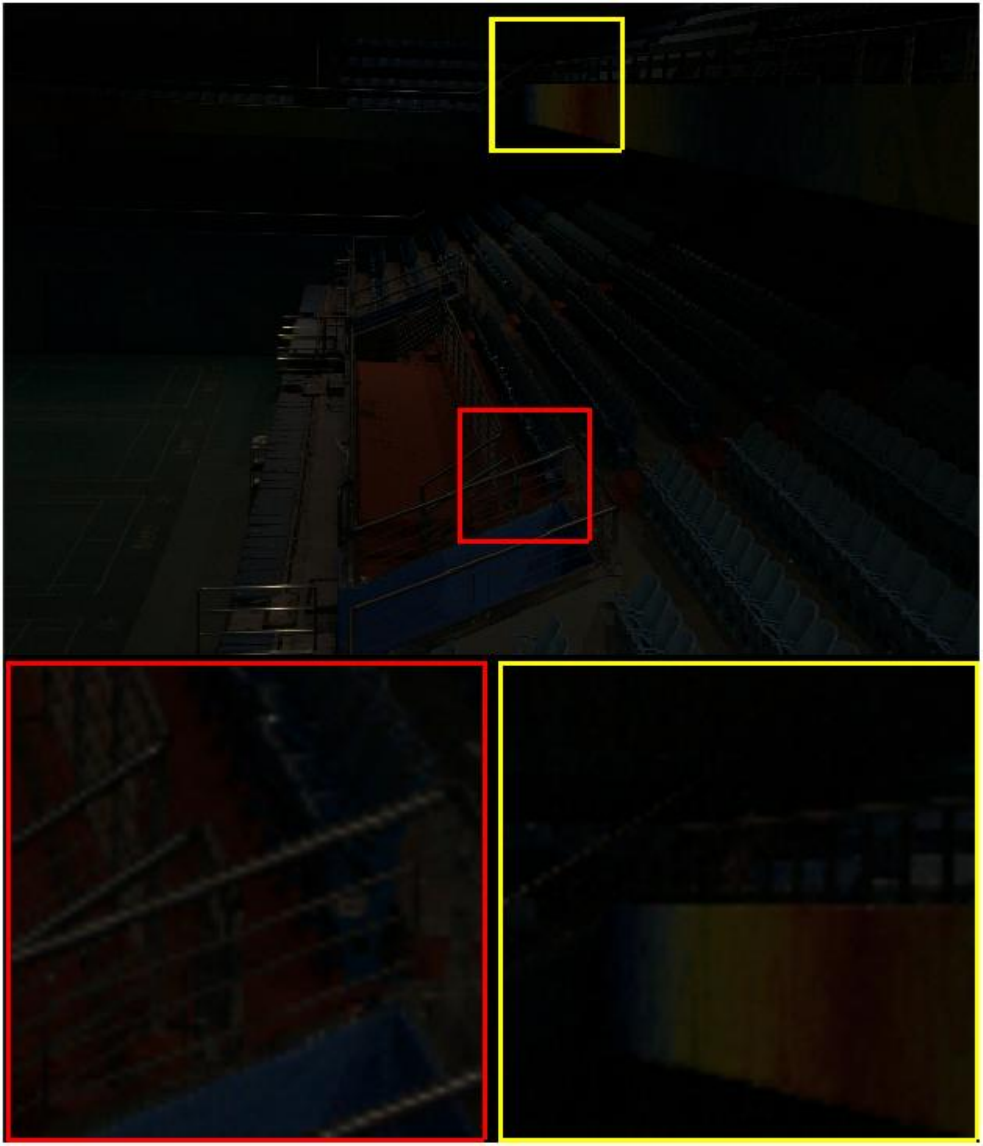}}
    \vspace{1pt}
    \centerline{\includegraphics[width=\textwidth]{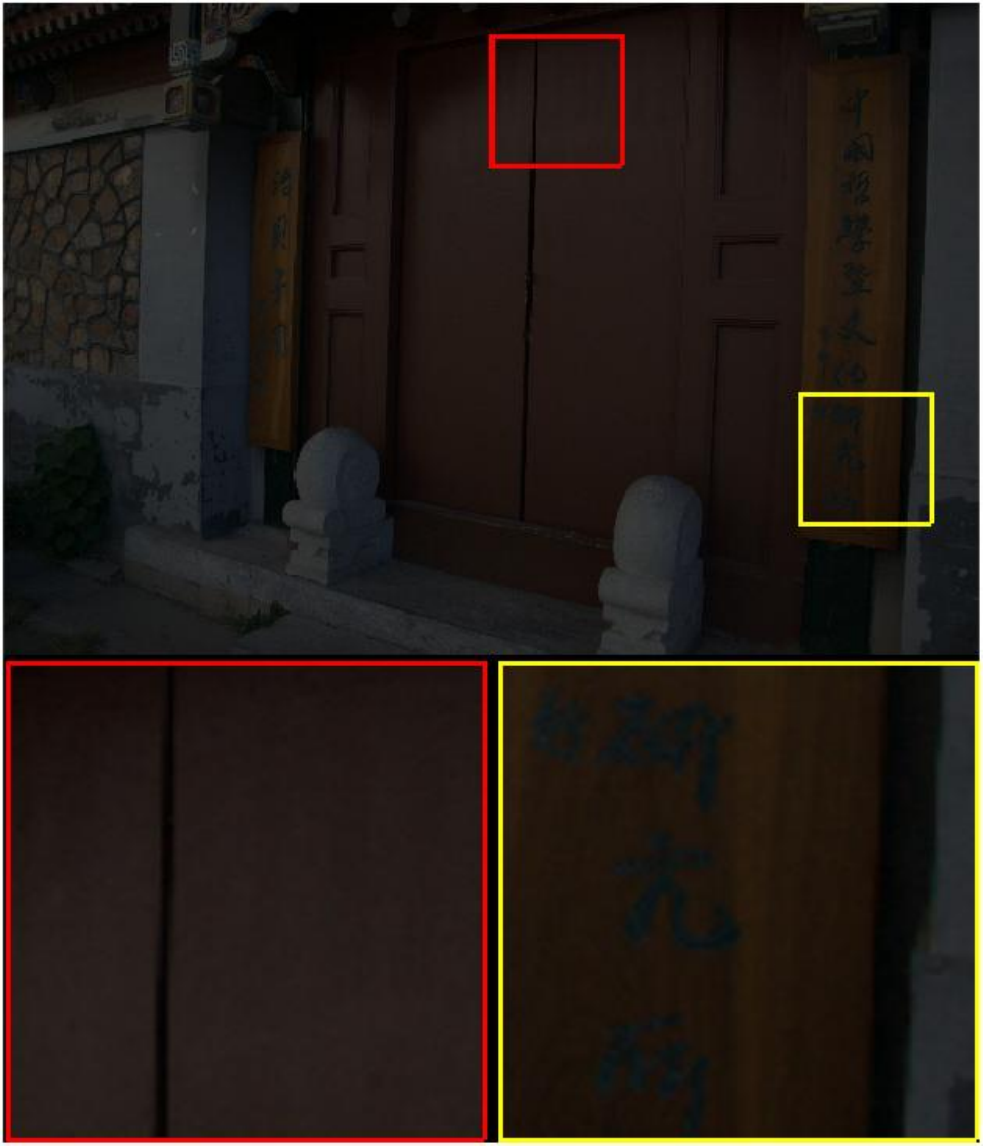}}
    \centerline{\footnotesize Input}
\end{minipage}\hfill%
\begin{minipage}[t]{\eightcolwidth}
    \centering
    \vspace{3pt}
    \centerline{\includegraphics[width=\textwidth]{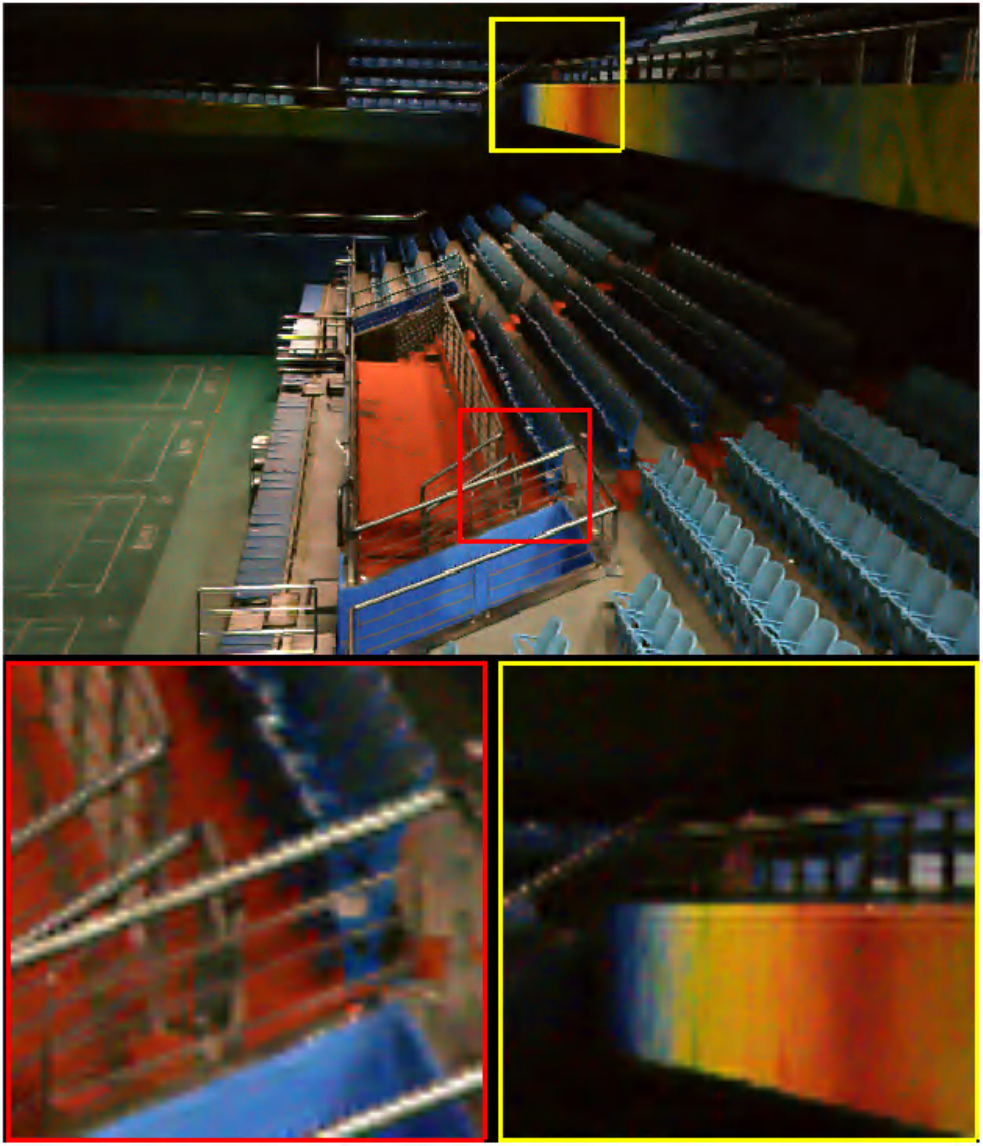}}
    \vspace{1pt}
    \centerline{\includegraphics[width=\textwidth]{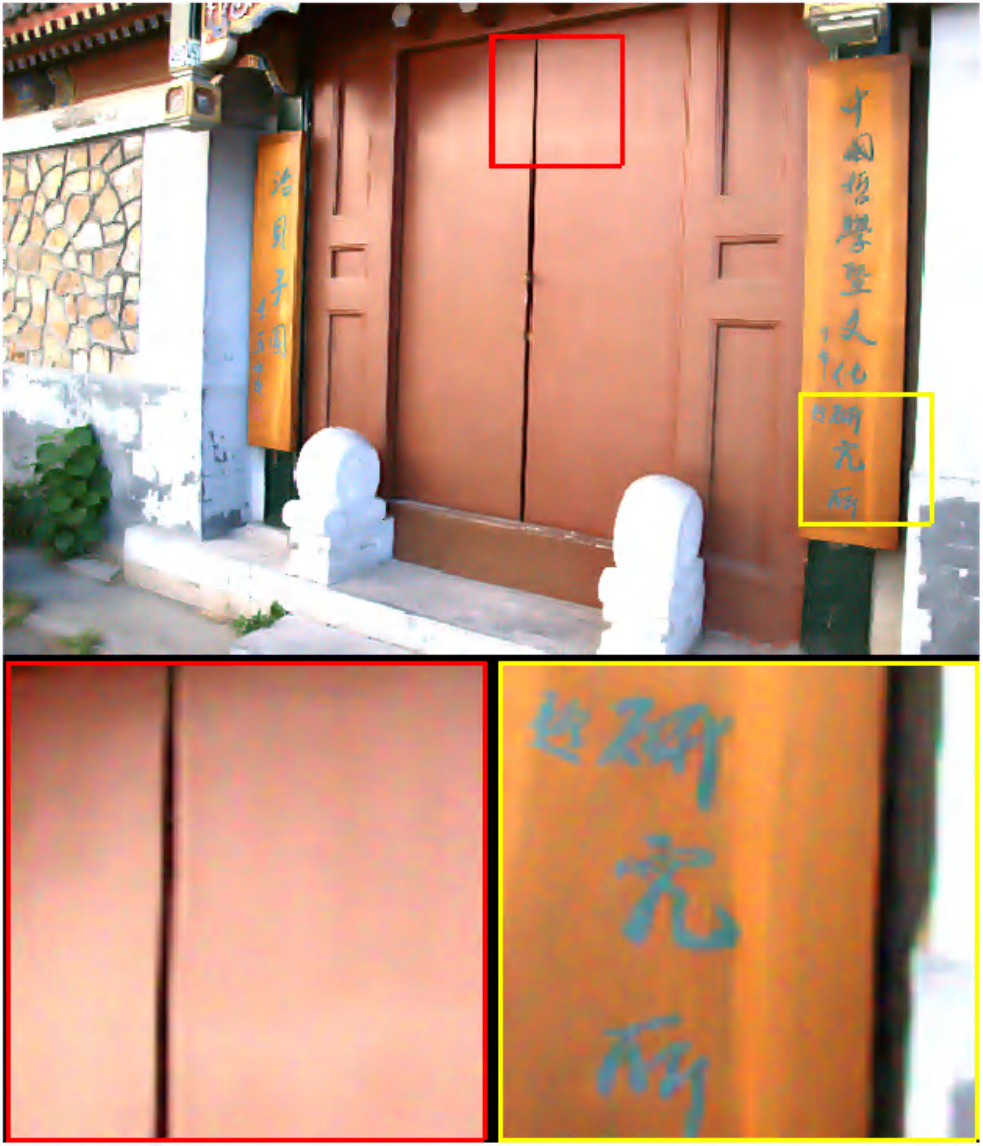}}
    \centerline{\footnotesize RUAS}
\end{minipage}\hfill%
\begin{minipage}[t]{\eightcolwidth}
    \centering
    \vspace{3pt}
    \centerline{\includegraphics[width=\textwidth]{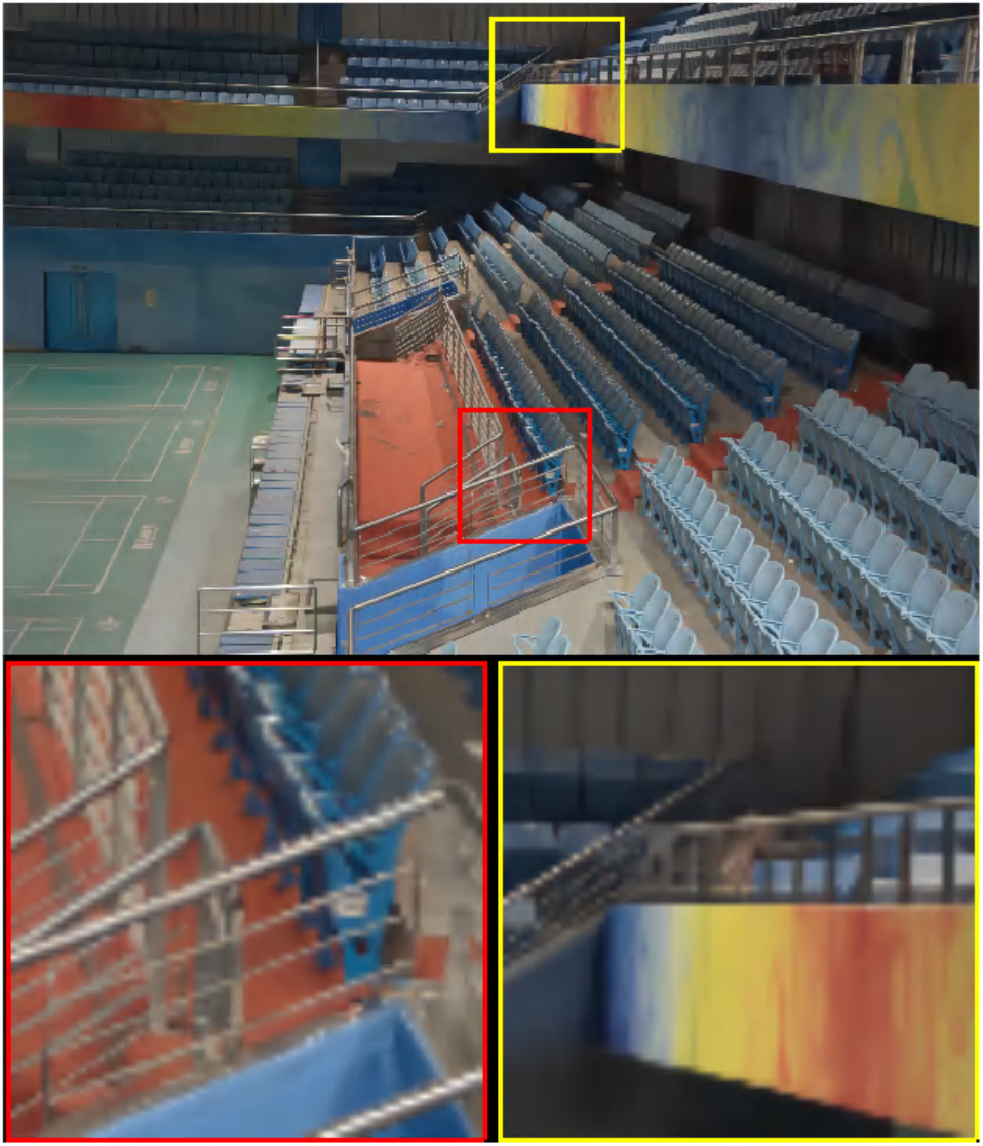}}
    \vspace{1pt}
    \centerline{\includegraphics[width=\textwidth]{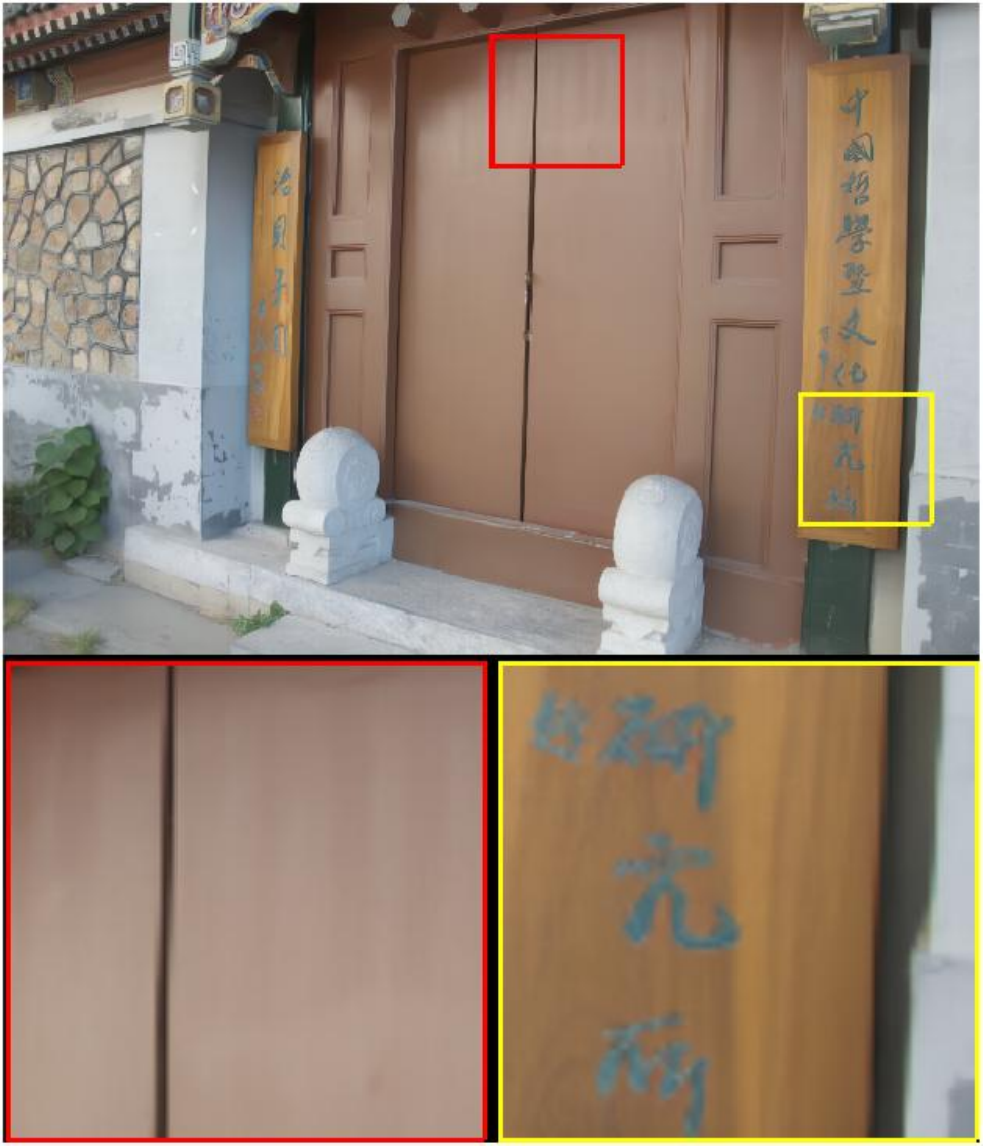}}
    \centerline{\footnotesize LLFlow}
\end{minipage}\hfill%
\begin{minipage}[t]{\eightcolwidth}
    \centering
    \vspace{3pt}
    \centerline{\includegraphics[width=\textwidth]{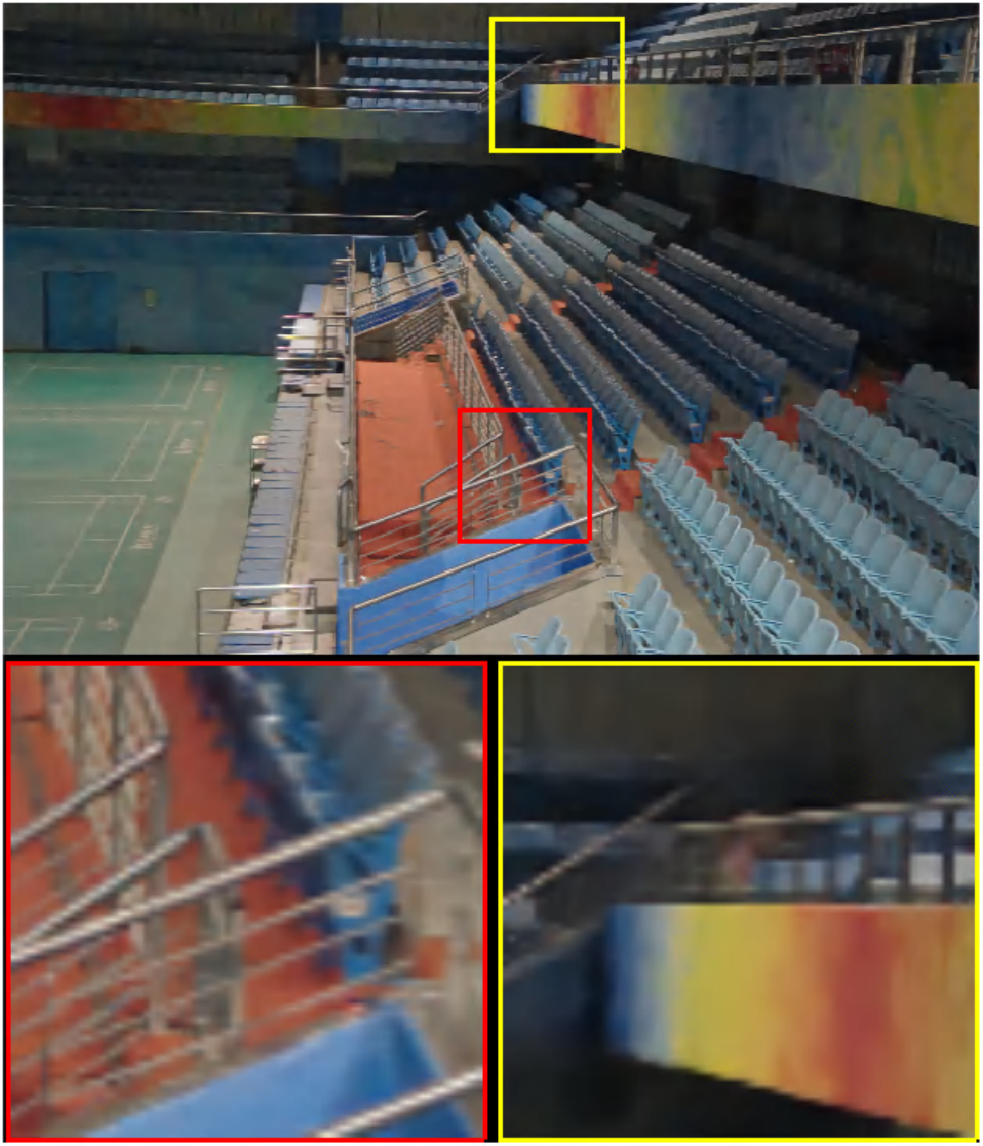}}
    \vspace{1pt}
    \centerline{\includegraphics[width=\textwidth]{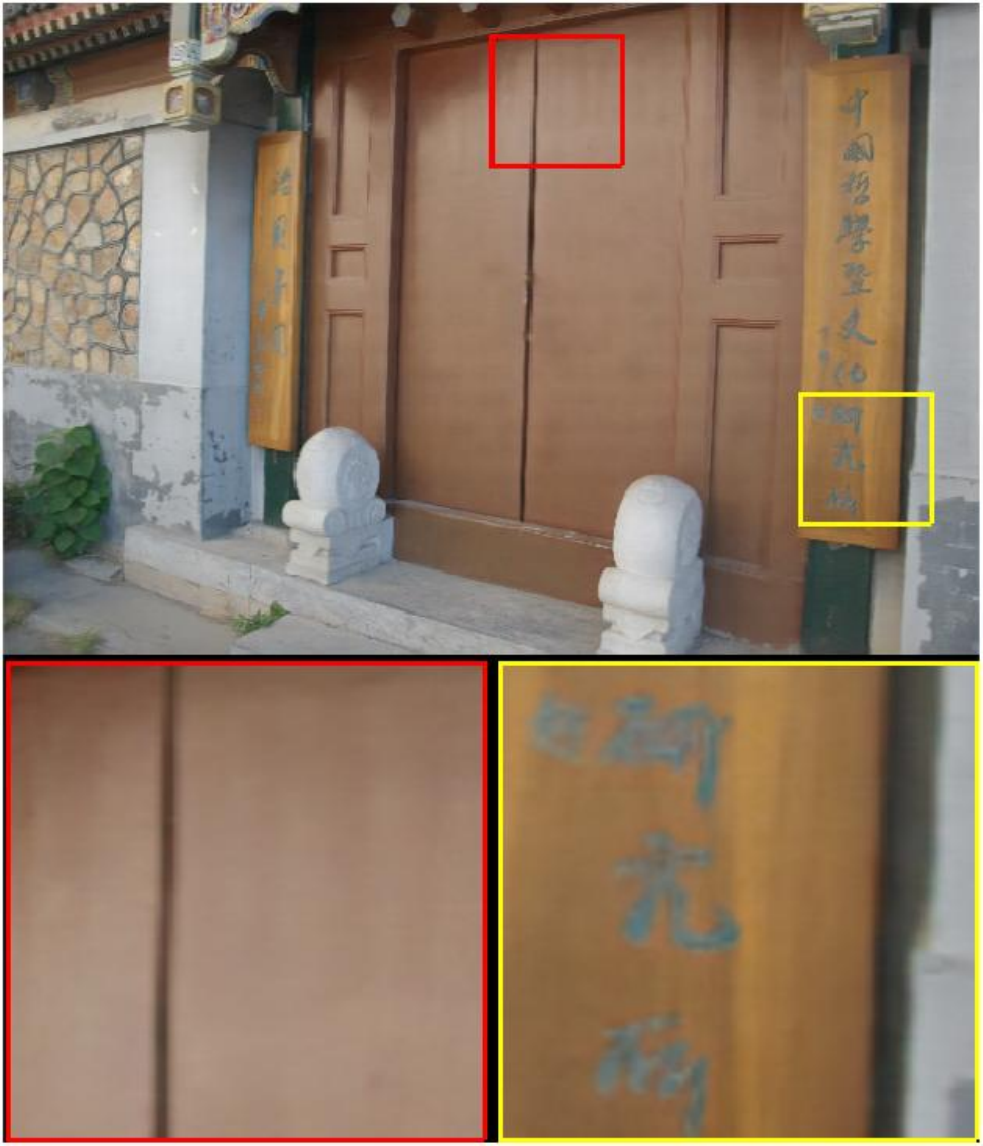}}
    \centerline{\footnotesize SNRNet}
\end{minipage}\hfill%
\begin{minipage}[t]{\eightcolwidth}
    \centering
    \vspace{3pt}
    \centerline{\includegraphics[width=\textwidth]{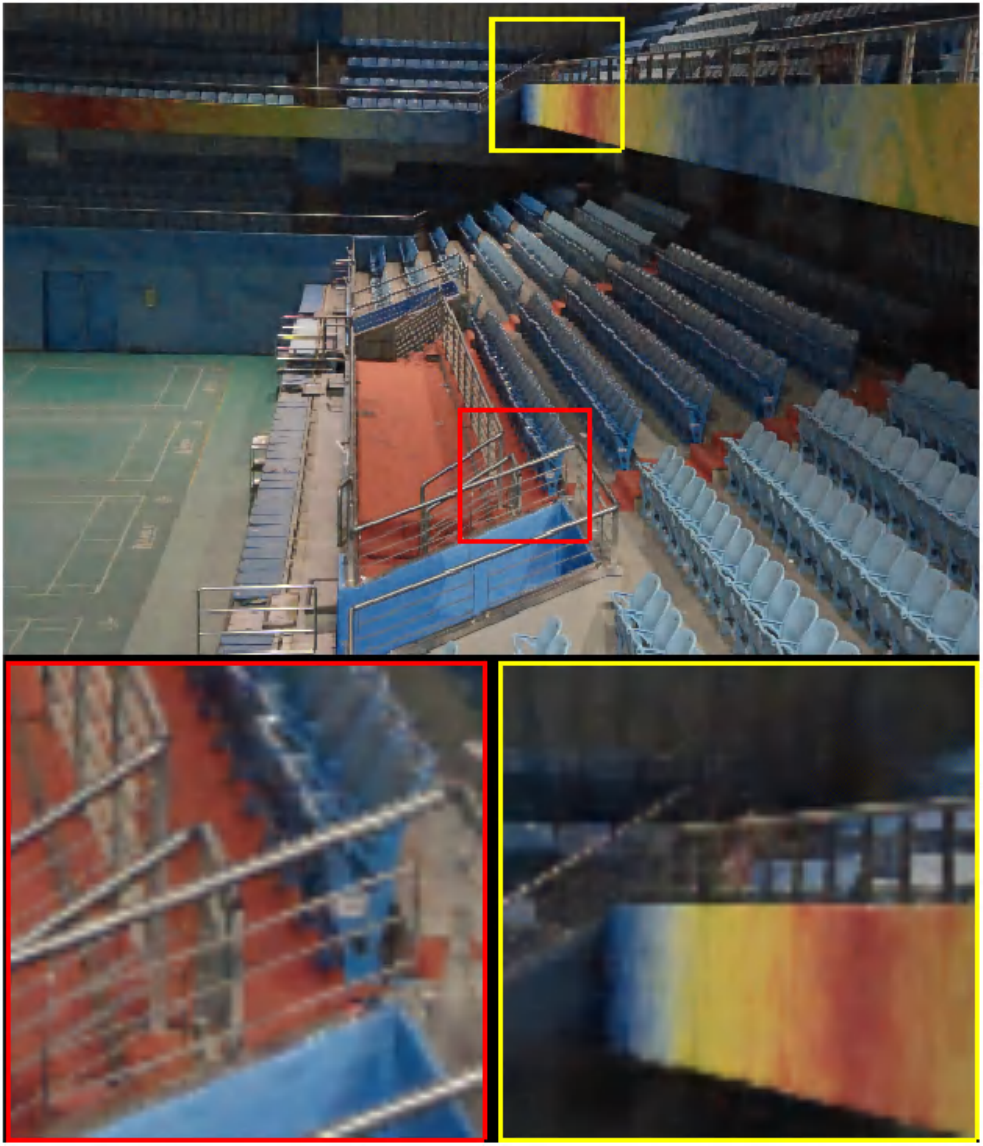}}
    \vspace{1pt}
    \centerline{\includegraphics[width=\textwidth]{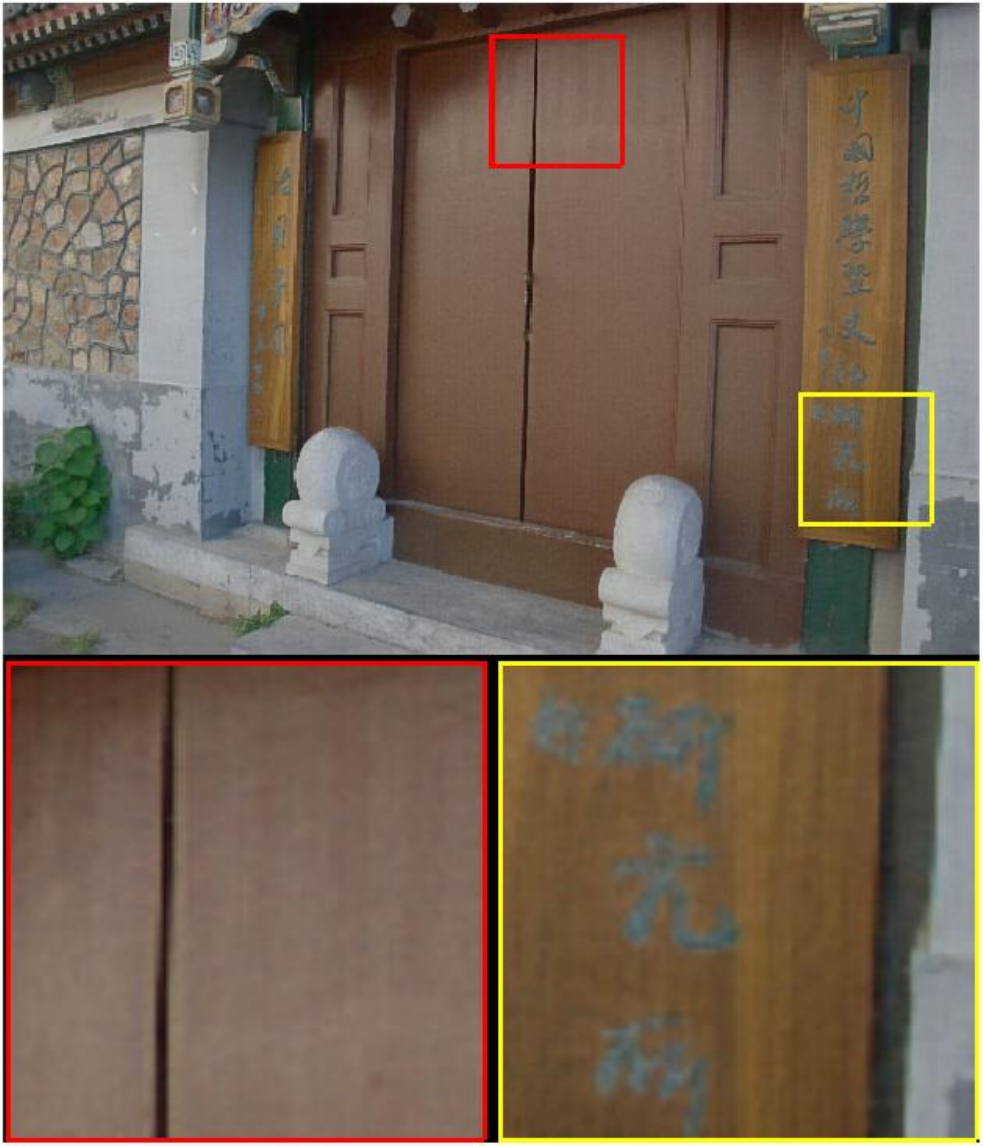}}
    \centerline{\footnotesize Retinexformer}
\end{minipage}\hfill%
\begin{minipage}[t]{\eightcolwidth}
    \centering
    \vspace{3pt}
    \centerline{\includegraphics[width=\textwidth]{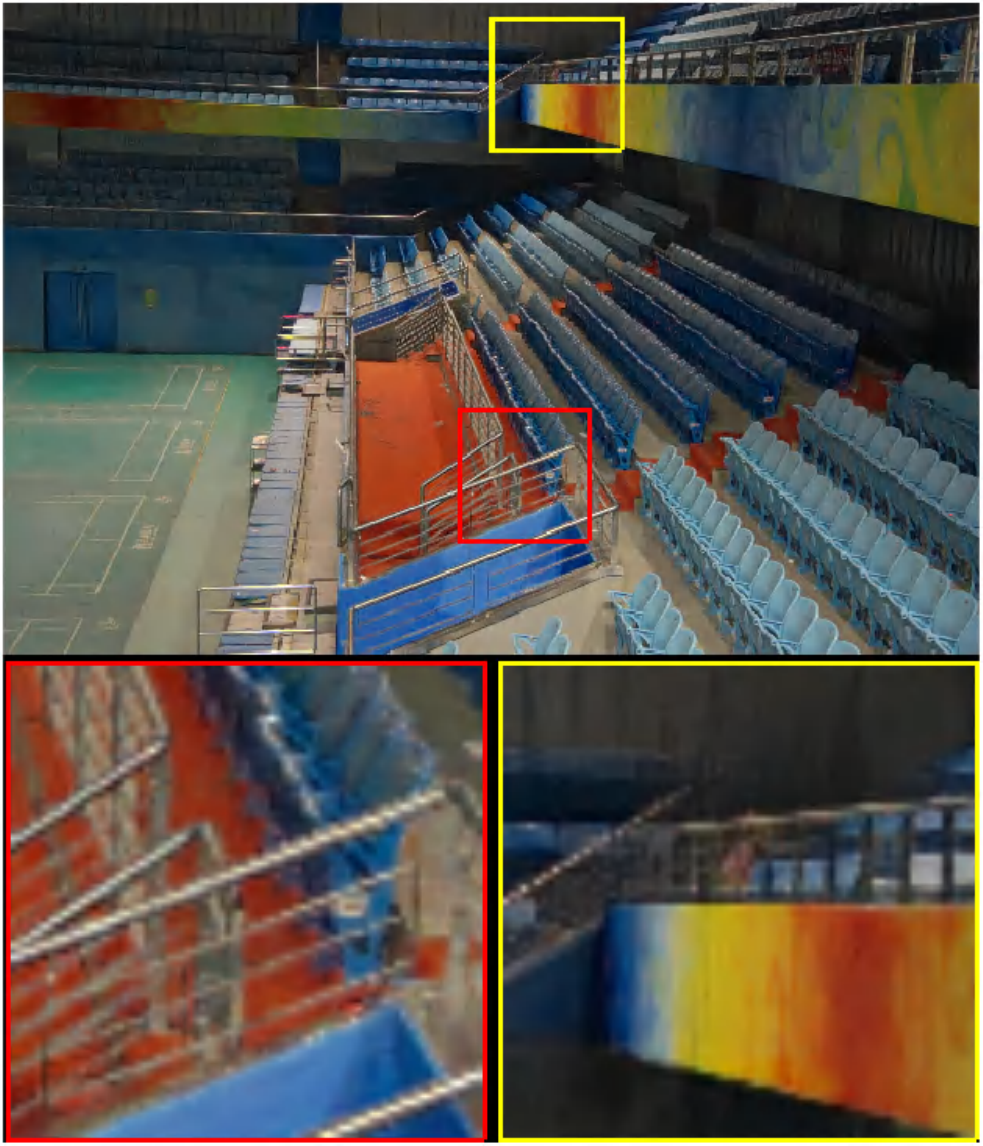}}
    \vspace{1pt}
    \centerline{\includegraphics[width=\textwidth]{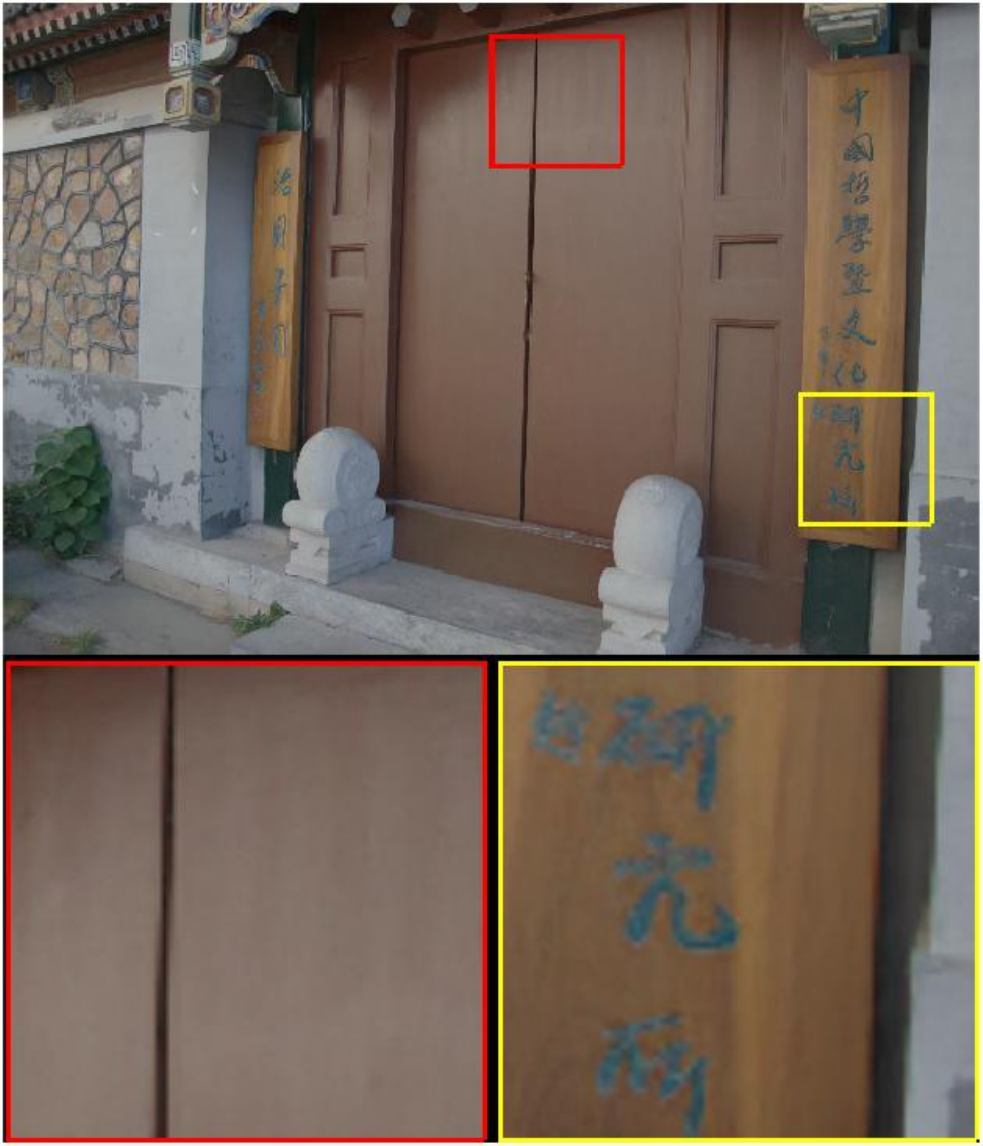}}
    \centerline{\footnotesize CIDNet}
\end{minipage}\hfill%
\begin{minipage}[t]{\eightcolwidth}
    \centering
    \vspace{3pt}
    \centerline{\includegraphics[width=\textwidth]{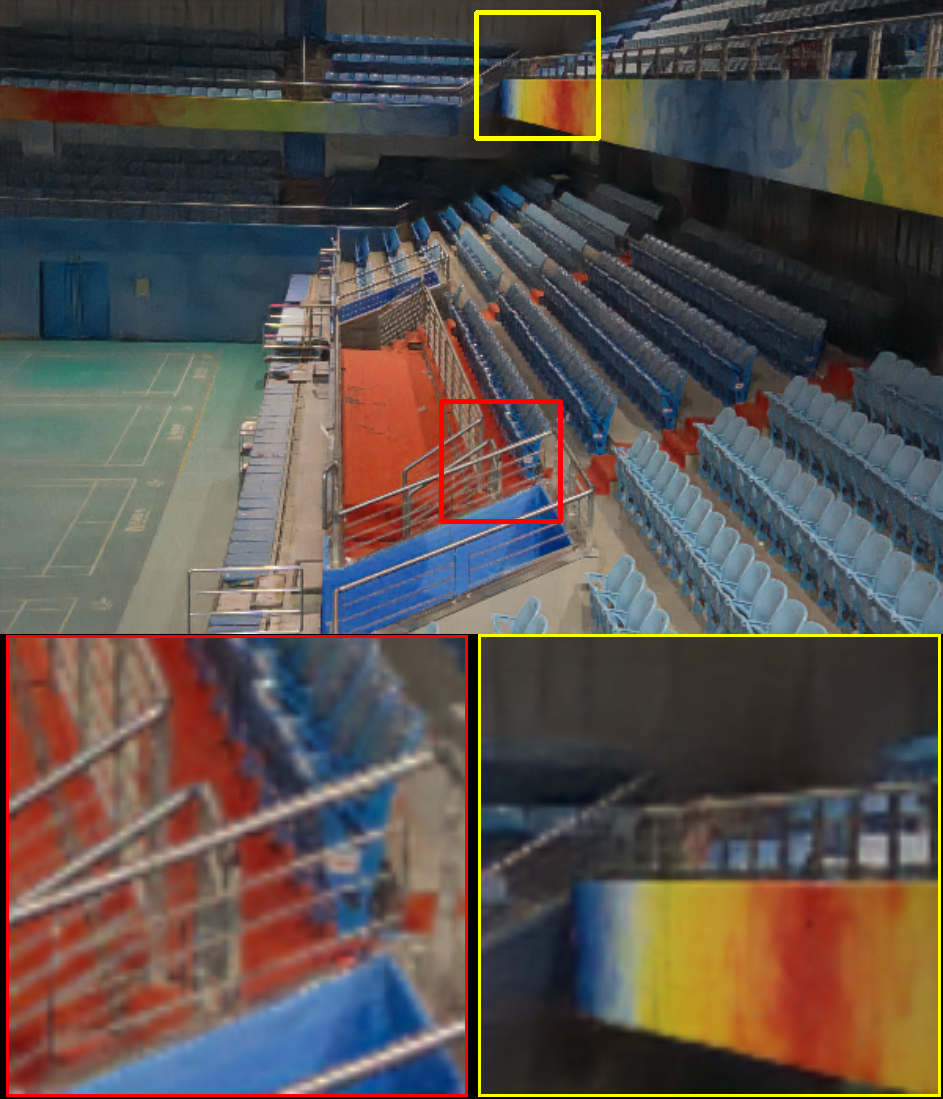}}
    \vspace{1pt}
    \centerline{\includegraphics[width=\textwidth]{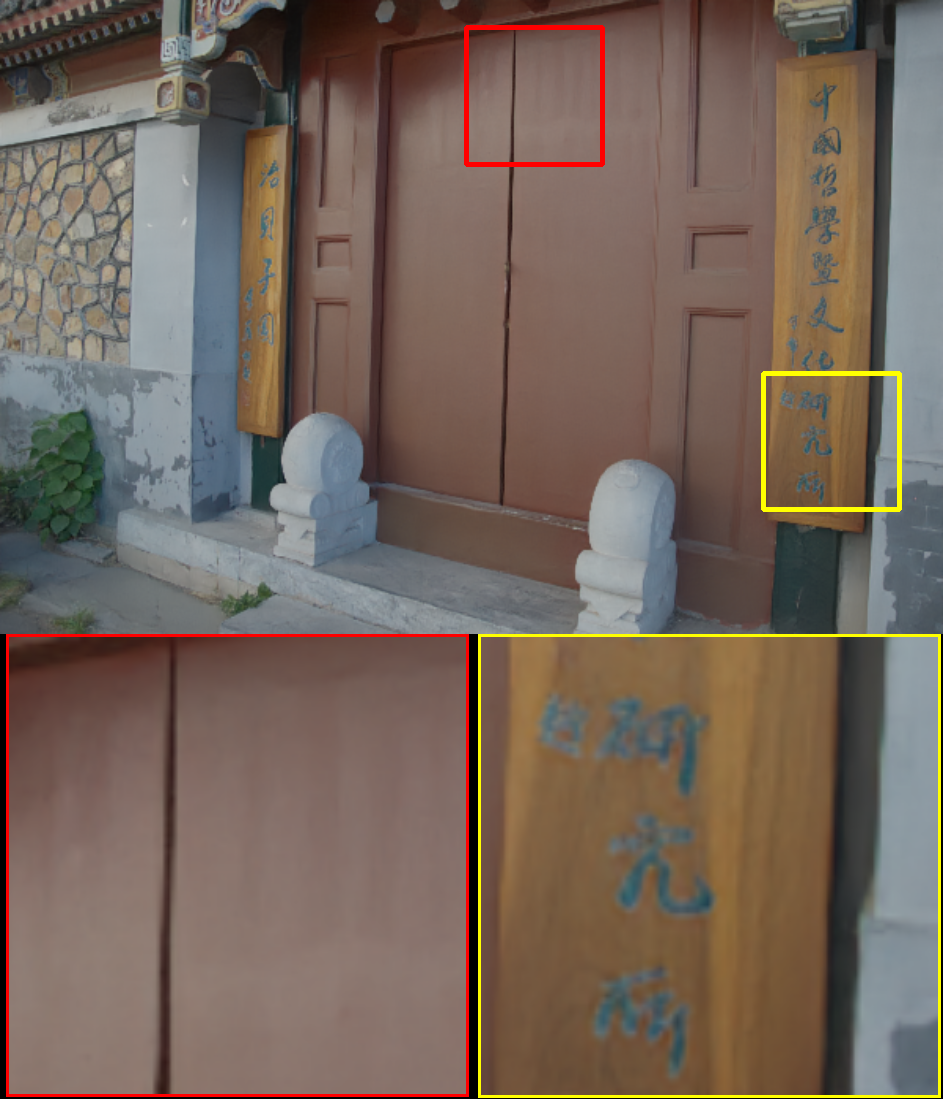}}
    \centerline{\footnotesize \textbf{Ours}}
\end{minipage}\hfill%
\begin{minipage}[t]{\eightcolwidth}
    \centering
    \vspace{3pt}
    \centerline{\includegraphics[width=\textwidth]{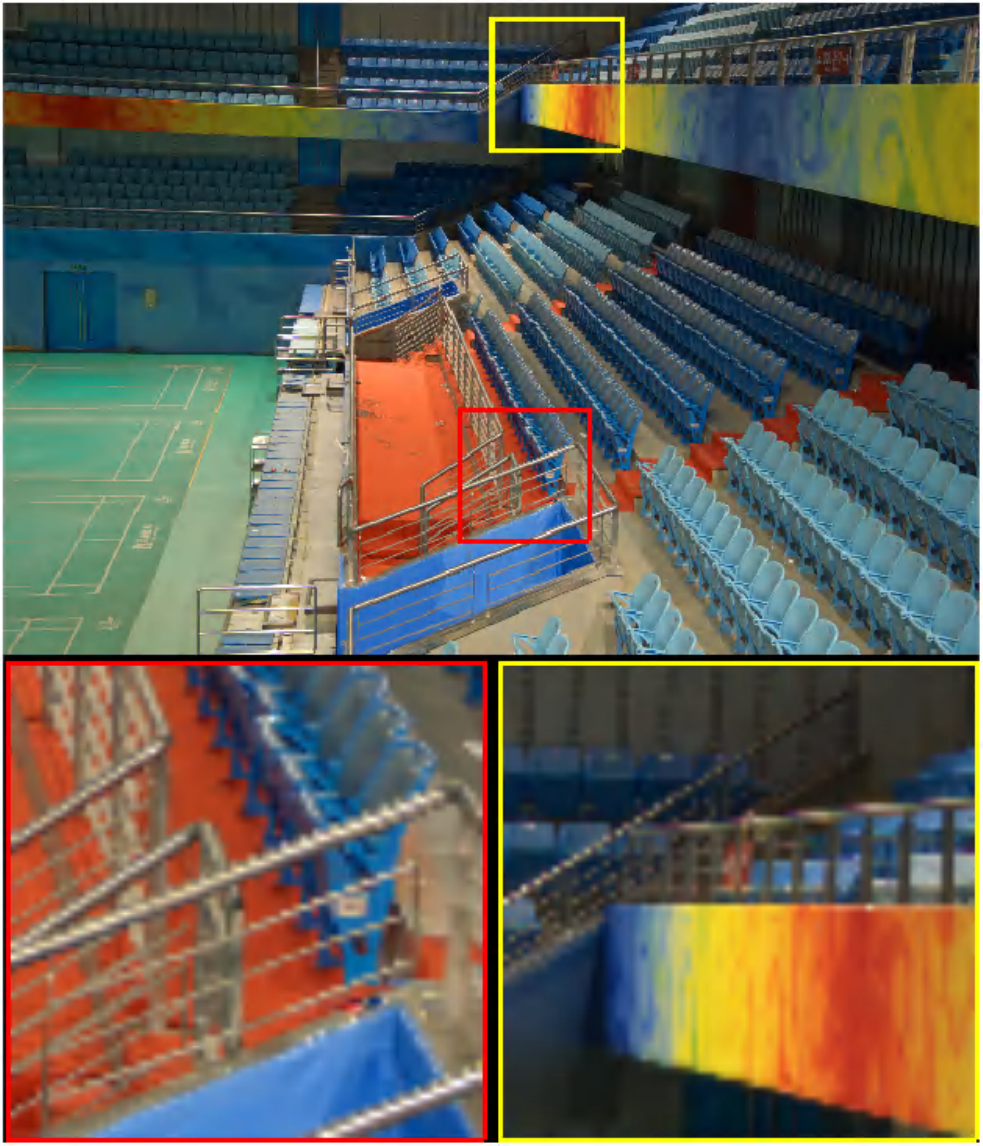}}
    \vspace{1pt}
    \centerline{\includegraphics[width=\textwidth]{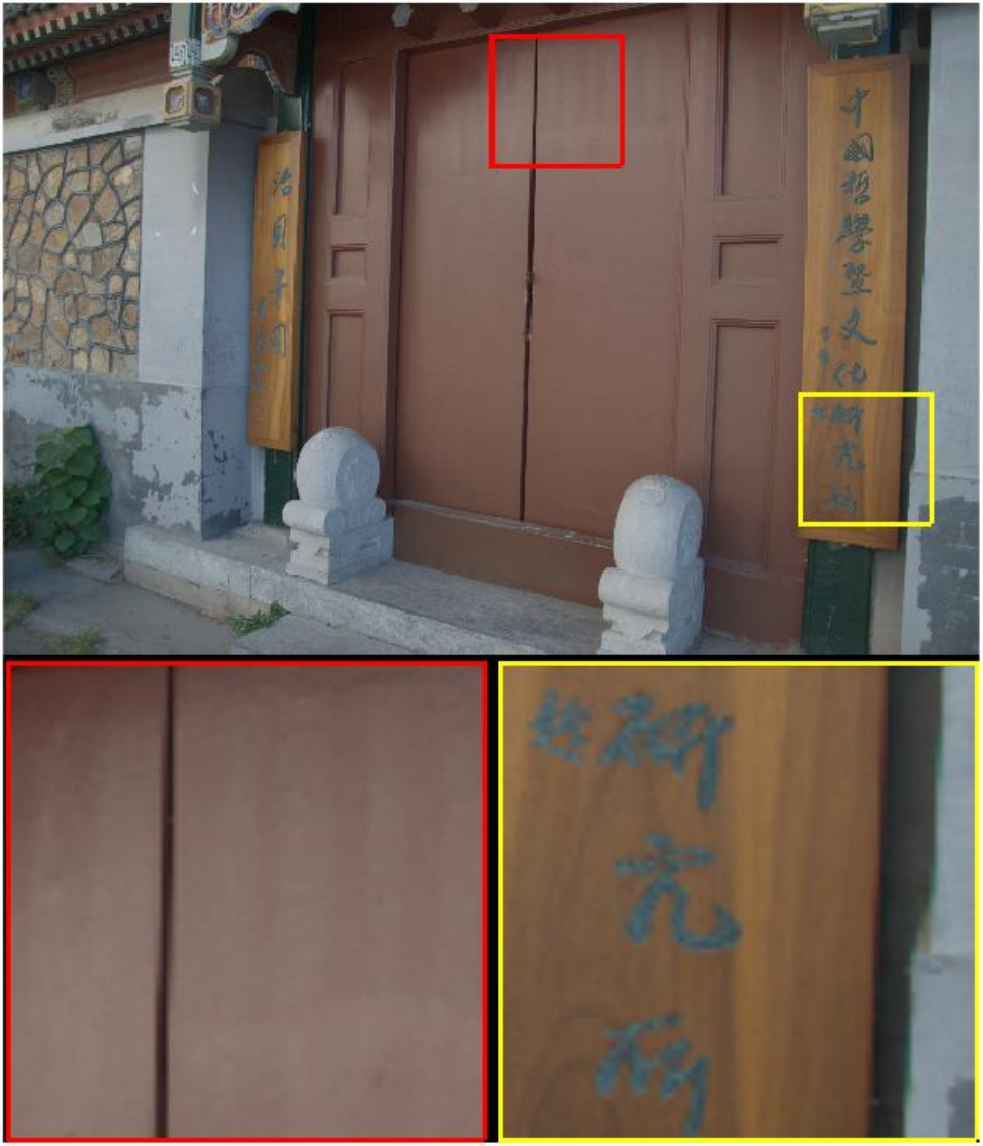}}
    \centerline{\footnotesize GroundTruth}
\end{minipage}

\caption{Visual comparison of enhanced results from SOTA methods on LOL-v1 (top) and LOL-v2-Real (bottom).}
\label{fig:LOL}
\end{figure*}

\begin{figure*}[!htb]
    \centering
    \includegraphics[width=1\linewidth]{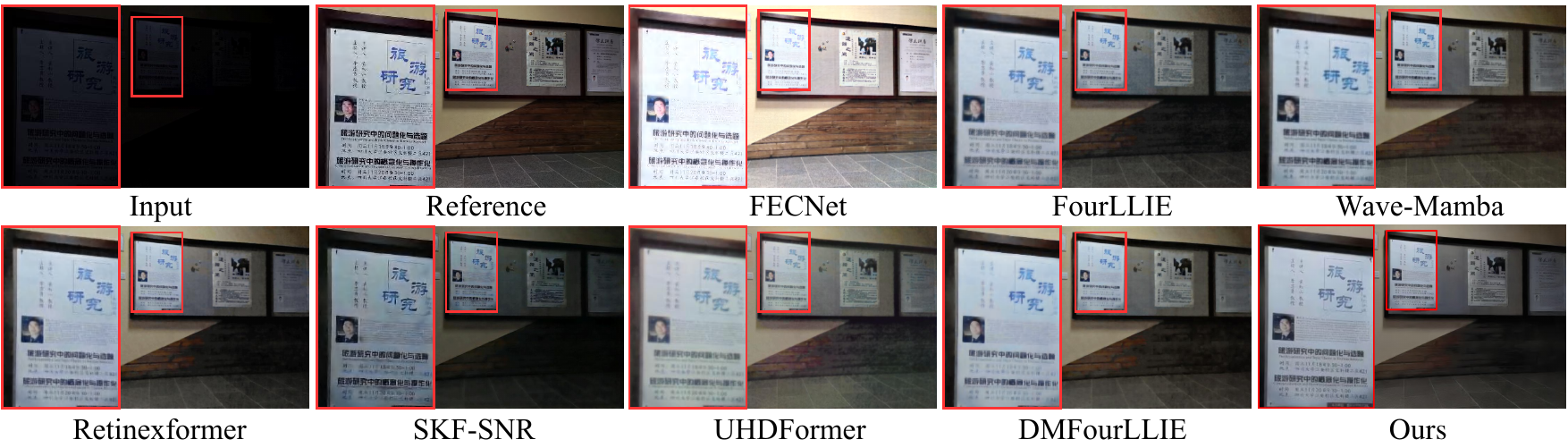}
    \caption{Visual comparison on LSRW-Huawei dataset.}
    \label{fig:huawei-com}
\end{figure*}

\begin{table}
    \centering
    \renewcommand{\arraystretch}{1.1}

    \vspace{1.8mm}

    \resizebox{\linewidth}{!}{
    \begin{tabular}{l|cc|cc}
        \toprule[1pt]
        \multirow{2}{*}{\textbf{Methods}} &
        \multicolumn{2}{c|}{\textbf{SICE}} & 
        \multicolumn{2}{c}{\textbf{SID}} \\
        \cline{2-5}
        ~ &
        PSNR$\uparrow$ & SSIM$\uparrow$ & 
        PSNR$\uparrow$ & SSIM$\uparrow$ \\
        \midrule
        RetinexNet \cite{LOL} &
        12.42 & 0.613 &
        15.70 & 0.395 \\
        ZeroDCE \cite{Zero-DCE} &
        12.45 & 0.639 & 
        14.09 & 0.090 \\
        RUAS \cite{RUAS} &
        8.66 & 0.494 & 
        12.62 & 0.081 \\
        URetinexNet \cite{URetinexNet} &
        10.90 & 0.605 & 
        15.52 & 0.323 \\
        LLFlow \cite{LLFlow} &
        12.74 & 0.617 & 
        16.23 & 0.367 \\
        CIDNet \cite{CIDNet} &
        13.44 & 0.642 &
        22.90 & 0.676 \\
        \textbf{Ours} &
        \color{red}{13.56} & \color{red}{0.653} &
        \color{red}{22.98} & \color{red}{0.681} \\
        \bottomrule[1pt]
    \end{tabular}
    }
        \caption{Quantitative result on SICE and Sony-Total-Dark datasets. The top-ranking score is in \textcolor{red}{red}.}
    \label{tab:SID}
\end{table}

\begin{table}
    \centering
    \renewcommand{\arraystretch}{1.1}

    \vspace{1.8mm}
    
    \resizebox{\linewidth}{!}{
        \begin{tabular}{l|c|cc}
        
            \toprule[1pt]
            
            \multirow{2}{*}{\textbf{Methods}} 
            & \multirow{2}{*}{\textbf{Venue}} 
            & \multicolumn{2}{c}{\textbf{LSRW-Huawei}} 
            \\

            \cline{3-4}

            & 
            & {PSNR}$\uparrow$ 
            & {SSIM}$\uparrow$ 
            \\

            \midrule
            KinD \cite{KinD} & MM'19 & 16.58 & 0.569 \\
            SNR-Aware \cite{SNR-Aware} & CVPR'22 & 20.67 & 0.591 \\
            Retinexformer \cite{Retinexformer} & ICCV'23 & \textcolor{blue}{21.23} & 0.631 \\
            FourLLIE \cite{FourLLIE} & MM'23 & 21.11 & 0.626 \\
            Wave-Mamba \cite{WaveMamba_24MM} & MM'24 & 21.19 & \textcolor{blue}{0.631} \\
            DMFourLLIE \cite{DMFourLLIE} & MM'24 & 21.09 & \textcolor{red}{0.633} \\
            DarkIR \cite{DarkIR} & CVPR'25 & 18.93 & 0.583 \\
            \textbf{Ours} & - & \textcolor{red}{21.39} & 0.625 \\
            \bottomrule[1pt]
        \end{tabular}
    }
        \caption{Quantitative result on LSRW-Huawei datasets. The top-ranking score is in \textcolor{red}{red} and the second best is in \textcolor{blue}{blue} color.}
    \label{tab:LSRW-Huawei}
\end{table}

\newcommand{\cmark}{{\color{green!70!black}\ding{51}}}
\newcommand{\xmark}{{\color{red!70!black}\ding{55}}}
\begin{table}[t]
\renewcommand{\arraystretch}{1.1}
\centering

\resizebox{\linewidth}{!}{
\begin{tabular}{c|ccc|cc|c}

\toprule
Index &ADPG & LGIM & ICDE & PSNR$\uparrow$ & SSIM$\uparrow$ & $\Delta$PSNR \\

\midrule

a &\xmark & \xmark & \xmark & 23.46 & 0.842 & -- \\

b &\cmark & \xmark & \xmark & 24.21 & 0.859 & +0.75 \\

c &\xmark & \cmark & \xmark & 24.27 & 0.859 & +0.81 \\

d &\xmark & \xmark & \cmark & 23.87 & 0.848 & +0.41 \\

\midrule

e &\cmark & \cmark & \xmark & 23.94 & 0.855 & +0.48 \\

f &\cmark & \xmark & \cmark & 24.32 & 0.856 & +0.86 \\

g &\xmark & \cmark & \cmark & 24.16 & 0.855 & +0.70 \\

\midrule

\bf Ours &\cmark & \cmark & \cmark & \textbf{24.78} & \textbf{0.866} & \textbf{+1.32} \\

\bottomrule

\end{tabular}}
\caption{Ablation study on the LOL-v1 dataset. We evaluate all combinations of three proposed components: ADPG, LGIM, and ICDE. $\Delta$ denotes the PSNR improvement over the vanilla baseline.}
\label{tab:ablation}
\end{table}

\subsection{Comparison with State-of-the-Art Methods}

We compare InterLight with a wide range of state-of-the-art methods, including Retinex-based methods (RetinexNet \cite{LOL}, KinD \cite{KinD}, RUAS \cite{RUAS}), RGB-based methods (Zero-DCE \cite{Zero-DCE}, EnlightenGAN \cite{EnGAN}, and LLFormer \cite{LLFormer}), and recent advanced models (Retinexformer \cite{Retinexformer}, DarkIR \cite{DarkIR}, CIDNet \cite{CIDNet}, and CWNet \cite{CWNet}).

\subsubsection{Results on LOL Datasets}
Table~\ref{tab:table-LOL} reports the quantitative results on LOL-v1 and LOL-v2 datasets. Our InterLight achieves consistent superiority across all subsets. On the LOL-v1 dataset, InterLight achieves the second-best PSNR of 24.78 dB and the best SSIM of 0.862. Notably, compared to the recent HVI-based method CIDNet, our method yields a significant improvement of 0.97 dB in PSNR, validating the effectiveness of our intrinsic learning paradigm over standard HVI transformations. On the LOL-v2-Real subset, InterLight ranks first with 24.06 dB PSNR and 0.866 SSIM, demonstrating robustness to complex real-world noise. On the LOL-v2-Synthetic, it also secures the top rank with 25.73 dB PSNR, significantly outperforming LightenDiff at much lower computational cost. The visual results in Figure \ref{fig:LOL} show that our method not only provides more accurate recovery of multi‑color regions than the CIDNet, but also delivers more stable brightness enhancement.

\subsubsection{Results on Results on SICE and SID Datasets}
To evaluate the capability of handling extreme degradations, we test on the SICE (multi-exposure) and SID (extreme low-light) datasets. As shown in Table~\ref{tab:SID}, InterLight consistently outperforms existing methods. Specifically, on the SID dataset, which presents extreme non-linear sensor noise and severe raw-to-RGB color shifts, our method achieves 22.98 dB PSNR, surpassing the strong competitor CIDNet by 0.08 dB and the flow-based LLFlow by over 6.7 dB. This proves our resilience against severe non-physical degradations, as the dynamic projection in the ADPG module effectively captures sample-specific non-linear prior contexts. Additionally, this highlights the advantage of our Luminance-Gated Intrinsic Memory (LGIM) in hallucinating valid details from extremely weak signals where local convolutions often fail. Similarly, on the SICE dataset, we achieve the best performance (13.56 dB), demonstrating robust dynamic range compression capabilities.

\subsubsection{Results on LSRW-Huawei dataset} 

We further evaluate InterLight on the LSRW-Huawei dataset, which involves images captured with sensor characteristics distinct from those in the LOL dataset. As shown in Table~\ref{tab:LSRW-Huawei}, InterLight achieves the best PSNR of 21.39 dB, surpassing recent approaches such as DMFourLLIE and DarkIR. These results confirm that our Physics-Guided Augmentation (PGA) strategy effectively adapts internal feature distributions to unseen sensor domains.
As shown in Figure \ref{fig:huawei-com}, while InterLight
does not surpass DMFourLLIE in the SSIM metric in
Table \ref{tab:LSRW-Huawei}, its enhancement of character regions appears more realistic. This may be attributed to our method’s stronger ability to recover fine details in dark areas.

\subsection{Ablation Study}

To validate the contribution of each component in InterLight, we conduct a comprehensive ablation study on the LOL-v1 dataset. We evaluate the impact of the Adaptive Degradation Prior Generation (ADPG) framework, Luminance-Gated Intrinsic Memory (LGIM), and Intrinsic-Consistent Data Expansion (ICDE).

As shown in Table~\ref{tab:ablation}, the baseline model yields a PSNR of 23.46 dB. Incorporating ADPG (Index b) brings a substantial gain of 0.75 dB, demonstrating that extracting image-specific degradation priors is crucial for adaptive restoration. Furthermore, adding LGIM alone (Index c) improves performance by 0.81 dB, which validates our hypothesis that retrieving high-quality features from an internal memory bank effectively compensates for information loss in dark regions. Finally, employing ICDE (Index d) leads to a 0.41 dB improvement, confirming that our physics-aware augmentation facilitates the learning of more robust features.

Finally, the full model, synergizing all three components, achieves the peak performance of 24.78 dB, with a cumulative improvement of 1.32 dB over the baseline.Crucially, while the ICDE strategy provides an orthogonal improvement (+0.41 dB), our architectural modules deliver the vast majority of the gains (ADPG and LGIM independently contribute +0.75 dB and +0.81 dB, respectively). This confirms that our intrinsic structural design is the definitive core contributor, and all proposed modules are complementary and collectively contribute to the state-of-the-art performance of InterLight.

\section{Conclusion and Future Work}
In this paper, we propose InterLight, a principled LLIE framework that deeply mines intrinsic illumination  priors at both the data and feature levels. Through Intrinsic‑Consistent Data Expansion, the model learns illumination‑invariant representations under physically plausible perturbations. The Adaptive Degradation Prior Generation module captures sample‑specific degradation characteristics via a learnable dictionary, producing prompts that guide chrominance restoration in a spatially adaptive manner. Luminance‑Gated Intrinsic Memory further retrieves structural and textural cues with luminance‑aware modulation, enabling stronger compensation in dark areas while maintaining fidelity in bright regions. Extensive experiments across diverse benchmarks validate the effectiveness of InterLight.

Despite its strong performance, the dual-branch architecture with LGIM introduces additional computational overhead, and the physics-guided augmentation assumes relatively linear degradation models. In future work, we plan to explore model compression for lightweight deployment and extend this paradigm to video low-light enhancement by leveraging temporal coherence.

\section*{Acknowledgments}
This work was supported by the National Natural Science Foundation of
China under Grant 62302105, and in part by Funding by Science and Technology Projects in Guangzhou under Grant 2025A04J3851.




\bibliographystyle{named}
\bibliography{ijcai26}

\end{document}